\documentclass[default, iicol]{sn-jnl}

\usepackage{url}
\usepackage{footnote}
\makesavenoteenv{tabular}
\makesavenoteenv{table}
\usepackage{epsfig}
\usepackage{graphicx}
\usepackage{overpic}
\usepackage{amsmath,amssymb} % define this before the line numbering.
\usepackage{booktabs} % for \Xhline
\usepackage{algorithm}
\usepackage{algorithmicx}
\usepackage{algpseudocode}
\usepackage{graphics}
\usepackage{threeparttable}
\usepackage{color}
\usepackage[normalem]{ulem}
\usepackage{multirow}
\usepackage{float}
\usepackage{amsfonts}
\usepackage{bm}
\usepackage{array}
\usepackage{pifont}
\usepackage{rotating}
\usepackage{textcomp}
\usepackage{contour}
\usepackage{makecell}

\usepackage{enumitem}
\usepackage{colortbl}
\usepackage{bbding}
\usepackage{bigstrut}

\usepackage{silence}
\newcommand{\figref}[1]{Fig.~\ref{#1}}
\newcommand{\tabref}[1]{Table~\ref{#1}}
\newcommand{\equref}[1]{Eq.~(\ref{#1})}
\newcommand{\secref}[1]{Sec.~\ref{#1}}

\graphicspath{{./Imgs/},{./Imgs/Samples/},{./Imgs/ErrorAnalysis/},{./Imgs/Visualization/}, {./Imgs/authors/}}
\DeclareGraphicsExtensions{.pdf,.jpg,.png}

\def\PCASA{GPC}
\def\netname{GAPNet}
\def\sArt{{state-of-the-art~}}

\def\ie{\emph{i.e.}}

\def\etal{{\em et al.~}}
\def\sArt{{state-of-the-art~}}

\newcommand{\para}[1]{\vspace{0.05in}\noindent\textbf{#1}\quad}

\ifdefined \GramaCheck
  \newcommand{\CheckRmv}[1]{}
  \renewcommand{\eqref}[1]{Equation 1}
  \renewcommand{\equref}[1]{Equation 1}
  \renewcommand{\figref}[1]{Figure 1}
  \renewcommand{\tabref}[1]{Table 1}
\else
  \newcommand{\CheckRmv}[1]{#1}
  \renewcommand{\eqref}[1]{Equation~(\ref{#1})}
\fi
\jyear{2025}%

\begin{document}

%%%%%%%%% TITLE
\title[GAPNet]{GAPNet: A Lightweight Framework for Image and Video Salient Object Detection via Granularity-Aware Paradigm}

\author[1]{\fnm{Yu-Huan} \sur{Wu}}

\author[1]{\fnm{Wei} \sur{Liu}}
\equalcont{Corresponding author}

\author[2]{\fnm{Zi-Xuan} \sur{Zhu}}

\author[1]{\fnm{Zizhou} \sur{Wang}}

\author[1]{\fnm{Yong} \sur{Liu}}

\author[1]{\fnm{Liangli} \sur{Zhen}}
\equalcont{Corresponding authors}

\affil[1]{\orgdiv{Institute of High Performance Computing (IHPC)}, \orgname{A*STAR}, \orgaddress{\city{Singapore} \postcode{138632}}}

\affil[2]{\orgdiv{VCIP, College of Computer Science}, \orgname{Nankai University}, \orgaddress{\city{Tianjin, China} \postcode{300350}}}

% \author{Yu-Huan Wu, Wei Liu, Shi-Chen Zhang, Zizhou Wang, Yong Liu, and Liangli Zhen
%   \thanks{Y.-H.~Wu, W.~Liu,  Z.~Wang, Y.~Liu, and L.~Zhen are with the Institute of High Performance Computing (IHPC), Agency for Science, Technology and Research (A*STAR), Singapore.}
%   \thanks{S.-C.~Zhang is with VCIP, College of Computer Science, Nankai University, China.}
% }

% \markboth{IEEE Transactions on Image Processing}
% {IEEE Transactions on Image Processing} % The editoral office will adjust this.

%\thispagestyle{empty}

%Adopting lightweight backbones can reduce the requirements for computational resources but the performance is significantly compromised.
% Unlike existing methods that follow the common U-Net style paradigm,
% this paper introduces a novel encoder-decoder network with granularity-aware connections (\netname) tailored to lightweight SOD.

%%%%%%%%% ABSTRACT
\abstract{
Recent salient object detection (SOD) models predominantly rely on heavyweight backbones, incurring substantial computational cost and hindering their practical application in various real-world settings, particularly on edge devices.
This paper presents \netname, a lightweight network built on the granularity-aware paradigm for both image and video SOD.
We assign saliency maps of different granularities to supervise the multi-scale decoder side-outputs:
coarse object locations for high-level outputs and fine-grained object boundaries for low-level outputs.
Specifically, our decoder is built with granularity-aware connections which fuse high-level features of low granularity and low-level features of high granularity, respectively.
To support these connections, we design granular pyramid convolution (GPC) and cross-scale attention (CSA) modules for efficient fusion of low-scale and high-scale features, respectively.
On top of the encoder, a self-attention module is built to learn global information, enabling accurate object localization with negligible computational cost. 
Unlike traditional U-Net-based approaches, our proposed method optimizes feature utilization and semantic interpretation while applying appropriate supervision at each processing stage.
Extensive experiments show that
the proposed method achieves a new \sArt performance among lightweight image and video SOD models. 
Code is available at \url{https://github.com/yuhuan-wu/GAPNet}.
}


\maketitle

%%%%%%%%% BODY TEXT
\section{Introduction}
Salient object detection (SOD) 
aims to detect the most salient region of interest in images 
by approximating the human visual system \cite{zhou2021rgb, han2018advanced}.
Accurate SOD can benefit a variety of vision tasks, including visual tracking \cite{wang2018salient, liu2024spatial}, semantic segmentation \cite{zhang2019pattern, liu2020leveraging}, image editing \cite{cheng2010repfinder}, medical imaging \cite{wu2021jcs}, and robot navigation \cite{craye2016environment}.
Early SOD methods relied on hand-crafted low-level features that captured object details and boundaries but lacked high-level semantics ~\cite{jiang2013salient}, resulting in suboptimal object localization.
%Early SOD methods mainly relied on hand-crafted low-level features, which are effective at identifying object details and %boundaries~\cite{jiang2013salient}. However, these methods often overlooked high-level semantic features, leading to suboptimal performance in accurately locating salient objects.

Recently, the performance of SOD tasks has been significantly improved by applying Convolutional Neural Networks (CNNs) 
that can learn low-level features at the bottom layers and high-level features at the top layers \cite{wang2021salient}. 
Current \sArt regular models 
\cite{liu2016dhsnet, zhang2017amulet, zhao2019egnet, liu2019simple, pang2020multi, zhou2020interactive, wu2022edn, zhuge2022salient}
made several significant successes in recent years.
These models primarily utilize established network architectures~\cite{simonyan2014very, he2016deep, wu2021p2t, liu2024vision, wu2025low}, which can extract very powerful pretrained features.
However, these models incur substantial computational overhead, hindering deployment on energy-constrained edge devices.
%The numbers of parameters of the above-mentioned models are between 26M and 162M, leading to high storage memory, 
%high FLOPs, and low FPS. For instance, MINet \cite{pang2020multi} with VGG-16 backbone \cite{simonyan2014very} has 162M parameters, 
%and 650M memory is required to store the model. 
%The FLOPs is above 87G and the FPS at batch size 30 is only 122 given an powerful NVIDIA RTX3090 GPU. 
%Such high requirements for computing resources make these heavyweight SOD models hard to deploy to devices with limited computing resources.

Notably, these constraints have sparked growing interest in lightweight SOD. 
% Representative lightweight models including HVPNet \cite{liu2020lightweight}, SAMNet \cite{liu2021samnet}, 
% and EDN-Lite \cite{wu2022edn} adopt simpler network structures and have much fewer parameters. 
% Taking EDN-Lite \cite{wu2022edn} as an example,
% it adopts MobileNet-V2 \cite{sandler2018mobilenetv2} as the backbone and has 1.8M parameters. 
% The FLOPs is only 0.75G, which is $99$\% smaller than the 87G FLOPs of the MINet \cite{pang2020multi}. 
% The FPS at batch size 30 of the former model is around ten times higher than that of the latter model given the same GPU. 
%However, the actual computational cost (\ie, number of FLOPs) and inference speed of these extremely lightweight models are not significantly reduced compared with lightweight models.
%On the other hand,
However, existing lightweight models, such as EDN-Lite~\cite{wu2022edn} and SAMNet~\cite{liu2021samnet}, face challenges in achieving comparable performance to heavyweight counterparts due to their use of lightweight backbones like  EfficientNet-B0 \cite{cheng2013efficient} and MobileNet-V2 \cite{sandler2018mobilenetv2}.
These backbones often compromise multi-level feature representation capabilities, leading to reduced accuracy.
To differentiate our work, we redesign the decoder to exploit the limited feature richness of lightweight backbones more effectively.
Instead of merely contrasting with heavyweight models, we show how our approach augments lightweight representations to narrow the performance gap.

%Compared to regular models, lightweight models have smaller memory requirements, lower computational cost, and higher inference speed. 
%However, these lightweight models adopt lightweight backbones like EfficientNet-B0 \cite{cheng2013efficient} and MobileNet-V2 \cite{sandler2018mobilenetv2}, which compromises their multi-level learning capabilities seen in larger networks like ResNet-50 \cite{he2016deep}, 
%leading to a notable decrease in performance. 
%Therefore, it is crucial to develop effective decoders that maintain efficiency without compromising performance.

We illustrate popular SOD decoders in \figref{fig:ed-archs}(a) and \figref{fig:ed-archs}(b).
Early methods \cite{li2016deep, hou2019deeply} (\figref{fig:ed-archs}(a)) use late fusion strategies, which directly conduct the prediction from the (fused) features from one or multiple stage(s).
These decoders are very efficient due to simple architectures, but come with less effective performance.
Recently, U-Net styles (\figref{fig:ed-archs} (b)) are more popular in SOD and have been adopted by many approaches \cite{liu2019simple, zhou2020interactive}.
Through top-down feature fusion with deep supervision, they delve into multi-scale low-level and high-level feature learning, which is essential to achieve high performance.
However, U-Net-based decoders are not specifically tailored for lightweight models, leading to inefficiencies in leveraging multi-level features and suboptimal performance when deployed on limited-resource platforms.
%However, they are less efficient and the semantics could be easily affected by low-level features through progressive top-down feature fusion. 
%Besides, applying same supervision to different stages is suboptimal, 
%not considering the feature characteristics for each stage.
%Another popular variant of U-Net is the cascaded partial decoder \cite{wu2019cascaded}, which effectively utilizes the high-level features in a cascaded manner. 
%Although these paradigms are not customized for lightweight models, most of which still apply these strategies as the cornerstone. 

%Recent works also approve that high-level feature learning is very helpful for SOD.

Based on the above observations,
we propose an encoder-decoder structure, as shown in \figref{fig:ed-archs}(c), with granularity-aware paradigm (\netname) tailored to lightweight SOD. 
First, we introduce \textbf{G}ranularity-\textbf{A}ware \textbf{C}onnections to refine the low-level and high-level features separately, which are supervised by the non-center and center ground-truths, respectively.
%the proposed granular-aware connections can enhance feature utilization and semantic interpretation at various stages.
Then, an efficient cross-scale global guidance is incorporated to ensure the accurate localization of salient objects at each fusion stage.
% For low-level outputs, we use high-granularity boundary information to supervise.
% On top of the backbone encoder, we build an efficient but effective transformer instead of traditional CNNs to better learn global dependencies which are crucial for SOD.
% These global features are then integrated with encoder features by the decoder.
To enable effective low-level feature fusion at the bottom side, a granular pyramid convolution module with attention refinement (\PCASA) is constructed to enhance global perception.
For high-level feature fusion, we build an efficient  cross-scale attention block (CSA) to replace traditional CNN modules.
Since the spatial dimensions of high-level features are very low, adopting an attention block in our lightweight model is computationally efficient.
Compared to other styles, our proposed granularity-aware connections more effectively optimize the utilization and semantic interpretation of features at each stage, as well as employing targeted supervisions to optimize the performance.

\CheckRmv{%
\begin{figure}[!t]
    \centering
    \includegraphics[width=\linewidth]{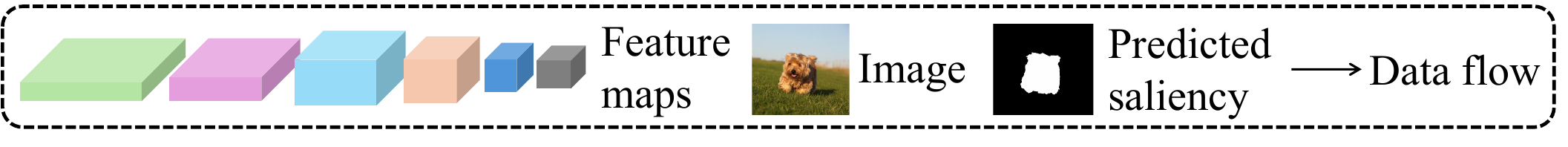} 
    % \vspace{0.1cm}
    \centering
    \resizebox{\linewidth}{!}{%
    \begin{tabular}{ccc}
        \includegraphics[height=50mm]{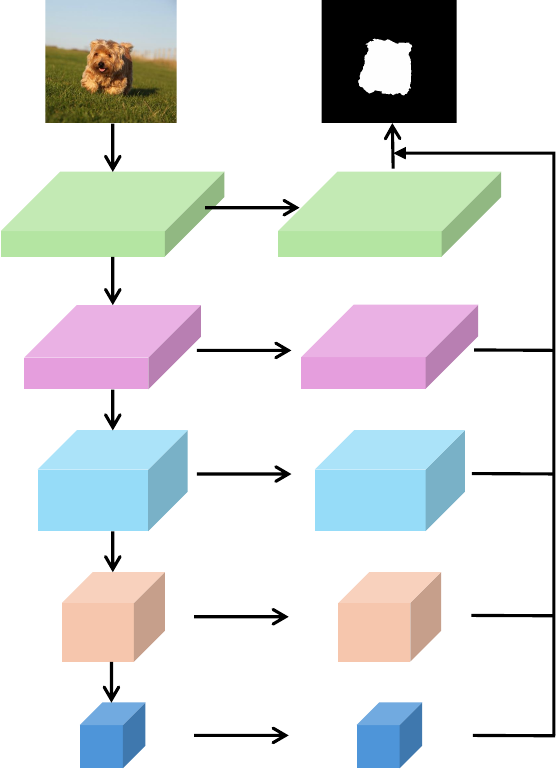} &
        \includegraphics[height=50mm]{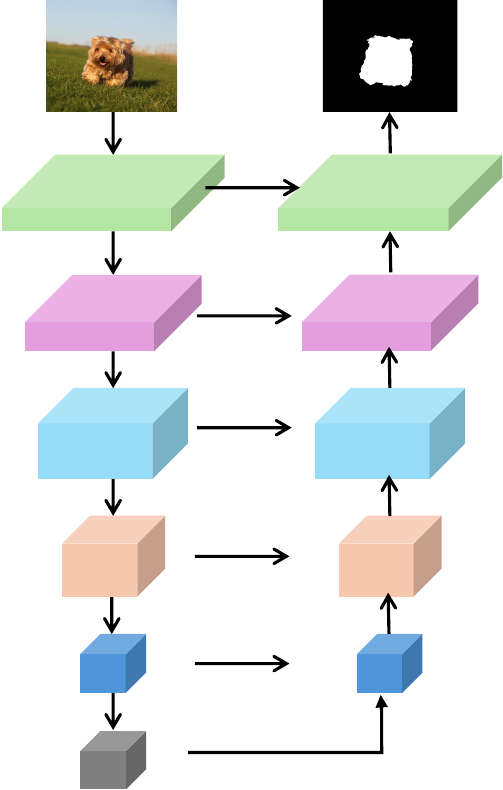} &
        \includegraphics[height=50mm]{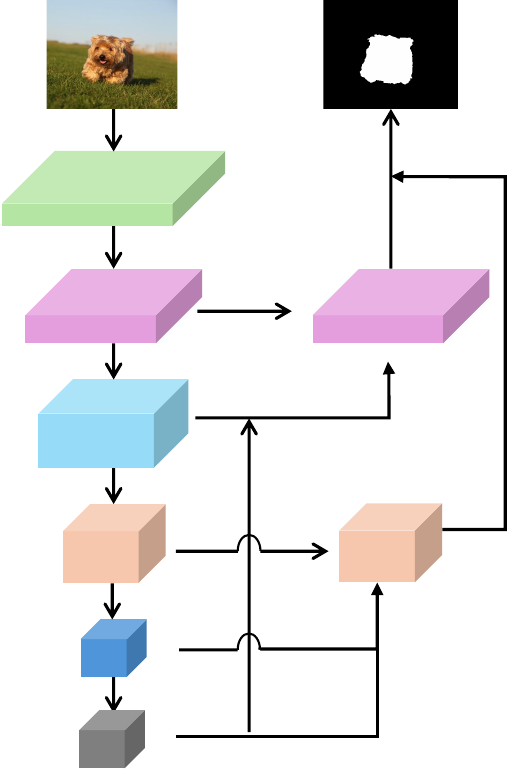}
        \\
        (a) Late Fusion %\cite{xie2015holistically, hou2019deeply} 
        &
        (b) U-Net %\cite{wu2022edn} 
        &
        % (c) CPD \cite{wu2019cascaded} &
        (c) Ours
    \end{tabular}
    }
    \caption{\textbf{Different encoder-decoder architectures}. 
(a) Late-fusion decoder side-output is calculated with corresponding encoder features only. Intermediate side-outputs are aggregated to generate the final output.
(b) U-Net decoder side-output is calculated with encoder features and higher-level decoder features or global features in a progressive top-down manner.
% (c) CPD is similar to U-Net but utilizes the high-level encoder features only.
(c) Ours GAPNet fuses global features with low-level and high-level encoder features to compute side-outputs which are then fused as the final output. The high-level and low-level side-outputs have low granularity and high granularity, respectively.
% (a) and (b) all use the full saliency map to supervise the side-outputs of different scales while
% (d) uses low-granularity center saliency to supervise high-level side-output and high-granularity boundary saliency to supervise low-level side-output. The final output is supervised by the full saliency map.
}
    \label{fig:ed-archs}
\end{figure}
}

The key novelty and main contributions of this paper are twofold:
%
% \begin{itemize}
% %
% \item We propose a novel lightweight encoder-decoder framework, {\netname}, that features granularity-aware connections. We design These connections fuse features at different scales with tailored supervisory signals. Specifically, coarse object-level guidance is applied to high-level features and fine-grained boundary guidance is applied to low-level features, resulting in improved feature utilization and semantic interpretation.
% \item To support granularity-aware connections, we design two efficient modules: the Granular Pyramid Convolution (GPC) for enhancing low-level features with multi-scale aggregation and attention refinement, and the Cross-Scale Attention (CSA) module for capturing global dependencies in high-level features. Together, these modules provide a balanced, efficient, and effective fusion process, significantly narrowing the performance gap compared to heavyweight models.
% %with potent global perception capabilities is built for the effective fusion of high-level features.
% \end{itemize}

\begin{itemize}
\item We introduce a granularity-aware paradigm for lightweight image/video SOD that couples scale-specific connections with matching supervision: high-level features learn from coarse object cues, while low-level features are guided by fine boundaries, yielding maximal feature reuse and coherent semantics throughout the pipeline.

\item We implement the paradigm with two compact fusion blocks: Granular Pyramid Convolution (GPC) that enriches low-level features via multi-scale aggregation and attention, and Cross-Scale Attention (CSA) that injects global context into high-level features. Coupled with a lightweight global-attention head, they narrow the accuracy gap to heavyweight models while remaining edge-friendly.
\end{itemize}

\section{Related Work} \label{sec:related}
Salient object detection (SOD) is one of the most significant tasks in computer vision, which can benefit many popular areas like visual tracking \cite{wang2018salient, liu2024spatial},, image editing \cite{cheng2010repfinder}, medical imaging \cite{liu2023revisiting, ji2022video, ji2024frontiers}, and camouflaged object detection \cite{fan2020camouflaged, ji2023deep}.
In the field of SOD, early popular methods were based on hand-crafted features \cite{liu2014superpixel, yang2013saliency, huang201950, li2013saliency, cheng2015global, wang2017salient}.
Deep-learning methods have since dominated SOD owing to their strong generalization across diverse scenarios.
The literature categorizes SOD methods into regular, lightweight, and extremely lightweight models.
We also review recent advances in encoder-decoder structures and multi-scale fusion.
At last, we introduce recent advances of video SOD.

\para{Regular models.} 
Traditional SOD models rely on complex network structures and usually require high computing resources for deployment.
The encoder-decoder structure has dominated SOD models where a heavy backbone is used to encode multi-scale features and a decoder is then deployed to fuse these features \cite{wu2021decomposition, wang2019salient, wang2019iterative, li2020stacked, yao2021boundary, wu2022edn, pang2020multi, wang2024feature, hao2025simple, pei2024calibnet}. On top of the encoder, some recent works \cite{chen2018reverse, zhao2021complementary, wu2022edn, yun2023towards, li2023rethinking} adopt additional CNN modules to extract global features to further improve the performance. 
% Zhao \etal \cite{zhao2021complementary} develops CTD with MobineNet-V2 backbone.
% The U-shape structure is decoupled into three branches which are then gradually merged with CAM and BRM to improve the region accuracy and boundary quality.
In general, heavyweight models achieve high detection accuracy at the cost of low model efficiency.

\para{Lightweight models.} 
Some works build lightweight SOD models with efficient feature fusion modules and lightweight backbones.
CSNet has only 100k parameters and is free of pre-training on ImageNet.
However, the estimation accuracy is not comparable to large models.
Liu \etal \cite{liu2020lightweight} proposed an efficient HVP module that emulates the primate visual cortex for hierarchical perception learning and builds HVPNet with 1.2M parameters.
SAMNet that encodes multi-scale features with a small network is developed in \cite{liu2021samnet}.
Fang \etal \cite{fang2022densely} presents lightweight DNTDF with EfficientNet-B0 backbone where PCSP is constructed to enhance the propagation of high-level features during decoding.
Wu \etal \cite{wu2022edn} proposed an extremely downsampled module on top of the encoder to extract global
features and build an effective decoder to recover object details from the global features.
The lightweight version EDN-Lite adopts MobileNet-V2 as the backbone and refreshes \sArt lightweight performance significantly.
% LARNet and LARNet* only have 0.66M and 0.09M parameters, respectively.
ADMNet \cite{zhou2024admnet} achieved near–heavyweight accuracy by fusing multi-scale context via a compact perception block and sharpening predictions with a dual-attention decoder.
Overall, lightweight models sacrifice detection accuracy for lower requirements for computing resources.

%\para{Extremely lightweight models.}
Recently,
some studies propose extremely lightweight SOD models,
%with even fewer parameters than , such as 
%CSNet \cite{cheng2021highly}, ELWNet \cite{wang2023elwnet}, and LARNet \cite{wang2023larnet}, 
exhibit several times fewer parameters than recent lightweight models.
For example,
CSNet \cite{cheng2021highly} introduced
a generalized OctConv block as the basic module for cross-stage multi-scale feature fusion.
In \cite{wang2023elwnet}, the wavelet transform fusion module (WTFM) is built by introducing the wavelet transform theory to CNNs
and then used to construct the extremely lightweight model ELWNet which has only 76K parameters.
Recently, LARNet and its variant LARNet* are built tailored to lightweight SOD \cite{wang2023larnet}.
The newly designed context gating module (CGM) proficiently enhances the features at all levels by transmitting global information.
Although the above methods are superior in terms of the model size, their FLOPs and throughput remain comparable to lightweight methods, and their accuracy still lags significantly behind.

\CheckRmv{%
\begin{figure*}[!t]
    \centering
    \includegraphics[width=\linewidth]{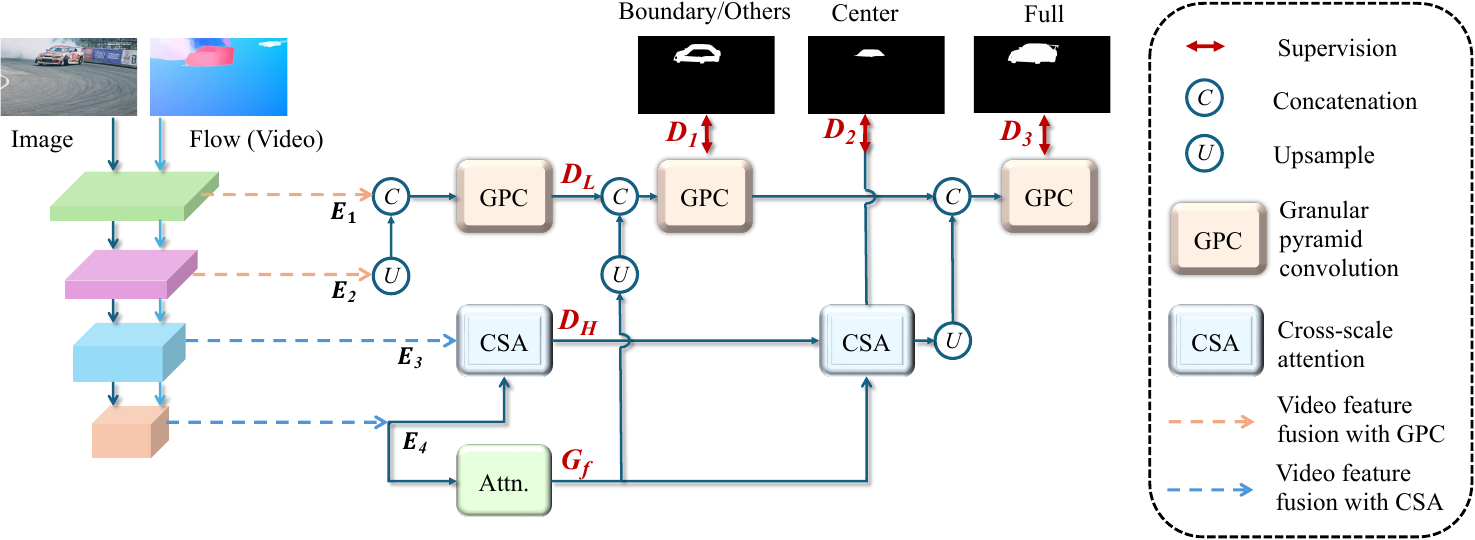}
    \caption{\textbf{Structure of the proposed network}.
    GT: ground truth.
    The first layer of backbone is not shown in this figure.
    {\PCASA} is used for fusion of low-scale features and CSA is used for fusion of high-level features.
    Low-scale side-output $D_1$ is supervised by boundary/others saliency of high granularity
    while high-scale side-output $D_2$ is supervised by center saliency of low granularity.
    Final output $D_3$ is supervised by the full saliency map.
    % Side-outputs $D_3$, $D_4$ and final output $D_5$ are supervised by the boundary-others saliency map, center saliency map and full saliency map, respectively.
    }
    \label{fig:our-network}
\end{figure*}%
}

\para{Encoder-decoder structures.} Many SOD models adopt the encoder-decoder structure to effectively learn multi-level multi-scale features \cite{ji2021cnn, wu2021regularized, chen2024collaborative, pei2022transformer, hao2025simple}. 
The encoder extracts features from the original image and the decoder integrates these features to the full saliency map using different manners.
The architectures include late fusion \cite{xie2015holistically, hou2017deeply}, its variant CPD \cite{wu2019cascaded}, U-Net \cite{wu2022edn, liu2021samnet} and its variants DNA \cite{liu2021dna} and CTD \cite{zhao2021complementary, li2023rethinking}.
The basic late fusion and U-Net architectures are illustrated in \figref{fig:ed-archs}(a) and \figref{fig:ed-archs}(b), respectively.
The former method generates the final output with late fusion and is more efficient due to its simple structure.
The latter method fuses features in a top-down manner and is more effective.
However,
the semantics could be easily affected by low-level features through
progressive top-down feature fusion. 
To take advantage of intermediate decoder features, a deep supervision mechanism \cite{lee2015deeply} has been applied to improve the performance of SOD.
% Existing works focus on designing novel architectures to generate side-outputs for deep supervision while ignoring to investigate which signals are effective to supervise different side-outputs.
% A variant of U-Net structure name CPD \cite{wu2019cascaded} is shown in \figref{fig:ed-archs}(c).
Existing works \cite{xie2015holistically, hou2017deeply, liu2021samnet, wu2022edn, liu2021dna} utilize the full saliency map to supervise side-outputs of different scales, introducing a significant performance improvement.
However, smaller features must be heavily upsampled, making uniform supervision suboptimal.
%Nevertheless, the side-outputs need to be upsampled to match the spatial resolution.
% The decoders of \cite{wu2022edn} and \cite{liu2021samnet} both have five stages, and the top-side feature resolution is only $\frac{1}{32}\times\frac{1}{32}$ of the full map.
% Hence, these features need to be upsampled $32\times32$ times
% by simple interpolation operation,
% making them less capable to represent fine-grained object boundary.
%Therefore, supervising smaller features with full saliency map can lead to suboptimal performance.

Although LDF \cite{wei2020label} contributed a label decoupling framework by decomposing saliency labels into body and detail maps,
this framework relies on iterative feature interactions and multiple training stages to refine predictions for heavyweight models.
Instead of iterative refinement, we introduce a granularity-aware supervision mechanism tailored to lightweight models within a single training stage. 
By directly assigning boundary-level guidance to low-level features and coarse object-level guidance to high-level features, our approach aligns supervision granularity with each decoder stage without relying on iterative feature interaction.

Moreover, most methods employ U-Net-based decoders \cite{wu2022edn, wei2020label, fan2020bbs, luo2024vscode, yin2025exploring}, which are not initially tailored for lightweight models. This leads to inefficiencies in leveraging multi-level features and suboptimal performance on resource-limited platforms.
Instead,
we propose a simpler and more direct decoder with granularity-aware connections that does not rely on complex iterations.
This design establishes a new lightweight-centric paradigm by matching each feature scale with appropriate supervisory signals. 
Our GPC and CSA modules, carefully devised for multi-scale feature fusion under lightweight constraints, 
help achieve balanced, effective, and resource-friendly SOD modeling on resource-limited devices.

\para{SOD in the video domain.}
In contrast to image-based SOD, video SOD generally incorporates the modeling of spatiotemporal features to capture both spatial appearance and temporal consistency across frames \cite{fan2019shifting, wang2020learning, chen2021exploring, ji2021full, gao2022weakly, cong2022psnet, qian2024controllable, chen2023dynamic}.
For example,
TENet \cite{ren2020tenet} employed the GT, the learnable prediction, and their weighted sum as an attention map. The weights gradually shift toward emphasizing the prediction as training progresses, thereby increasing the segmentation difficulty and improving spatial feature learning.
FSNet \cite{ji2021full} introduced a cross-attention which is computed between motion features and appearance features, enabling effective feature fusion that is subsequently used for salient object prediction.
DCFNet \cite{zhang2021dynamic} proposed to leverage the two adjacent frames of the current frame as temporal attention to guide information propagation. By employing matrix multiplication, it diffuses contextual cues throughout the entire spatial domain, achieving a dynamic filtering strategy with an effectively enlarged receptive field.
MMN \cite{zhao2024motion} applied two neighboring frames of the current frame as the memory to guide the extraction of high-level semantic features. 
This facilitates the integration of temporal information across frames, thereby enhancing the model's ability to accurately identify salient object characteristics.
Liu \etal \cite{liu2024learning} proposed using optical flow to guide the sampling window positions within input video clips, 
enabling more effective modeling of the spatial-temporal features of the same object.
Li \etal \cite{li2024novel} grouped keyframes based on background similarity and employed different models to learn each group. Each model focused on a specific type of background, 
thereby reducing the difficulty of modeling videos with frequent viewpoint changes.
Despite the above success, these works are with significant computational cost. Instead, we introduce a lightweight solution that is several times faster than existing heavyweight models and narrows the gap between lightweight and heavyweight models in video SOD.

\section{Methodology}
In this section, we first provide the details of our network structure in \secref{sec:structure}.
Then, we present our granularity-aware connections for multi-scale feature fusion in \secref{sec:featurefusion}.
Last, we introduce the granularity-aware deep supervision in \secref{sec:framework}.

\subsection{Network Structure} \label{sec:structure}

\figref{fig:our-network} presents the overall pipeline, which comprises an encoder (\secref{sec:encoder}), a global-feature extractor (\secref{sec:gfe}), and a decoder (\secref{sec:encoder}).
%We illustrate the overall pipeline in \figref{fig:our-network}, from which we can see that the proposed network consists of an encoder (\secref{sec:encoder}), 
%a global feature extractor (\secref{sec:gfe}) and a decoder (\secref{sec:decoder}).

\subsubsection{Backbone encoder}
\label{sec:encoder}
Due to computational constraints, we employ the well-known MobileNet-V2 \cite{sandler2018mobilenetv2} as the backbone.
%The MobileNet-V2 backbone is pretrained on ImageNet.
Following previous studies \cite{liu2020picanet, wu2022edn}, we remove the final pooling and fully connected layers to obtain a fully convolutional network suited to dense prediction.
The MobileNet-V2 encoder consists of five stages, with strides of 2, 4, 8, 16, and 32, respectively. The last four stages, denoted as \(E_1\), \(E_2\), \(E_3\), and \(E_4\), are utilized for decoding in our work.
These encoder features correspond to scales of \(\frac{1}{4}\), \(\frac{1}{8}\), \(\frac{1}{16}\), and \(\frac{1}{32}\), respectively. For simplicity, the first stage of the encoder is not depicted in the structure.
Our framework naturally extends to video sequences by incorporating temporal information through a two-stream architecture. For video inputs, we process both RGB frames and optical flow through separate lightweight backbones, fusing them at multiple hierarchical levels within our granularity-aware connections thereafter.

\subsubsection{Global feature extractor}
\label{sec:gfe}
As mentioned previously, the scale of the final encoder outputs is only \(\frac{1}{32}\) of the original input image.
Incorporating a global feature extractor with vision transformers is efficient at such a small scale. 
%making demanding attention blocks possible to implement.
Therefore, we stack a transformer module atop the encoder to extract global features, which are subsequently combined with local features for multi-scale feature fusion.
In the following sections, we will detail the global feature extractor.

% Firstly, pyramid feature maps with the depth-wise convolution are calculated and then concatenated as
% The global features are effective complement to local features to improve the performance of SOD.
% The global features have the same scale as the final encoder output $E_4$ ($\frac{1}{32}$ of the original map). 
% \begin{equation}
% G_f = \mathcal{F}(E_4)
% \end{equation}
% where $\mathcal{F}(\cdot)$ denotes the global feature extractor and $E_4$ is the output of the last encoder stage.

% \CheckRmv{
% \begin{equation}
% \begin{aligned}
% P_G^i &= {\rm AvgPool}(E_4), \quad i\in \{1,2,...,n\} \\
% P_{enc}^i &= {\rm DWConv}(P_G^i) + P_G^i \\
% P_G &= {\rm Concat}(P_{enc}^1, P_{enc}^2, ..., P_{enc}^n)
% \end{aligned}
% \end{equation}
% }
% where ${P_1, P_2, ..., P_n}$ are the pyramid feature maps and $n$ is the number of pooling layers.
% $P_n^{enc}$ is with feature maps after relative positional encoding.
% ${\rm DWConv}(\cdot)$ denotes depth-wise convolution with the kernel size $3\times3$.
% $\rm LayerNorm$ is performed after the concatenation.

Firstly, the attention is calculated as:
\CheckRmv{
\begin{equation}
\begin{aligned}
    % (Q, K, V) &= E_4(W^Q, W^K, W^V) \\
    % Att_G &= {\rm softmax}(\frac{QK^T}{\sqrt{d_k}})V \\
    Att_G &= E_4 + {\rm Attention}({\rm LayerNorm}(E_4)) \\
    % G_f &= LayerNorm(Att+FFN(Att))
\end{aligned}
\end{equation}
}
where $\rm LayerNorm(\cdot)$ denotes layer normalization. $\rm Attention(\cdot)$ is the self-attention defined as below:
\CheckRmv{
\begin{equation} \label{equ:van_attn}
\begin{aligned}
    (Q, K, V) &= X (W^Q, W^K, W^V) \\
    {\rm Attention}(X) &= {\rm Linear}({\rm softmax}(\frac{QK^T}{\sqrt{d_k}})V)
\end{aligned}
\end{equation}
}
where the input features are flattened with the spatial dimension, $\rm Linear(\cdot)$ denotes one linear transformation layer, $d_k$ is the scaling factor of the attention.

Then, an inverted residual block (IRB) \cite{sandler2018mobilenetv2} is applied as the feed-forward network (FFN) to compute the global features $G_f$, formulated as
\CheckRmv{
\begin{equation}
\begin{aligned}
    G_f &= Att_G + {\rm IRB}({\rm LayerNorm}(Att_G))
\end{aligned}
\end{equation}
}

% The dimension of pyramid feature map $P_G$ is much smaller than original input features $E_4$ by performing above pyramid pooling.
% Therefore, the efficient attention has much smaller computational complexity than vanilla attention. 
% Please refer to \cite{wu2022p2t} for more details on P2T block.

\subsubsection{Decoder network}
\label{sec:decoder}
In our \netname, the hierarchical decoder incorporates five feature fusion modules.
To maintain the efficiency of our framework, we have developed two types of modules for feature fusion in the decoder: granular pyramid pooling convolution (\(\text{\PCASA}\)) and cross-scale attention (CSA).
These modules are designed to fuse low-level and high-level features, respectively.
We will provide further details on these modules in \secref{sec:featurefusion}.
As depicted in \figref{fig:our-network}, low-level encoder features \(E_1\) and \(E_2\) are decoded to \(D_L\), and high-level encoder features \(E_3\) and \(E_4\) are decoded to \(D_H\), as calculated below:
\CheckRmv{
\begin{equation} \label{equ:dhl}
\begin{aligned}
    D_L &= \mathcal{H}{_P}({\rm Concat}({\rm Upsample}(\mathcal{G}(E_2)), \mathcal{G}(E_1))) \\
    D_H &= \mathcal{H}{_C}({\rm Concat}(\mathcal{G}(E_3), \mathcal{G}(E_4))),
\end{aligned}
\end{equation}
}
where $\mathcal{H}{_P}(\cdot)$ and $\mathcal{H}{_C}(\cdot)$ are the  {\PCASA} and CSA modules, respectively.
$\mathcal{G}(\cdot)$ denotes a convolution
followed by batch normalization and ReLU activation.
${\rm Upsample}(\cdot)$ upsamples low-scale features to the same resolution as high-scale features using bilinear interpolation.
For the concatenation of $\mathcal{H}{_C}(\cdot)$, it is not necessary to upsample the low-scale features because the spatial features are flattened into a vector before the concatenation.

Then, the decoder features $D_L$ and $D_H$ are fused with the global features $G_f$ to calculate low-level side-output $D_1$ and high-level side-output $D_2$, expressed as
\CheckRmv{
\begin{equation} \label{equ:d12}
\begin{aligned}
    D_1 &= \mathcal{H_P}({\rm Concat}({\rm Upsample}(G_f), D_L)) \\
    D_2 &= \mathcal{H_C}({\rm Concat}(D_H, G_f)),
\end{aligned}
\end{equation}
}

Last, the final decoder output is computed by fusing the side-outputs $D_1$ and $D_2$, shown as
\begin{equation} \label{equ:d3}
D_3 = \mathcal{H_P}({\rm Concat}({\rm Upsample}(D_2), D_1)),
\end{equation}
we employ GPC for the final fusion because it excels at preserving fine-grained boundary details at high spatial resolutions, which is essential for accurate final predictions. Additionally, GPC is computationally more efficient than CSA when processing the high-resolution concatenated features.

\CheckRmv{%
\begin{figure}[!t]
    % \centering
    % \begin{subfigure}{\linewidth}
    \centering
    \small
    \includegraphics[width=0.9\linewidth]{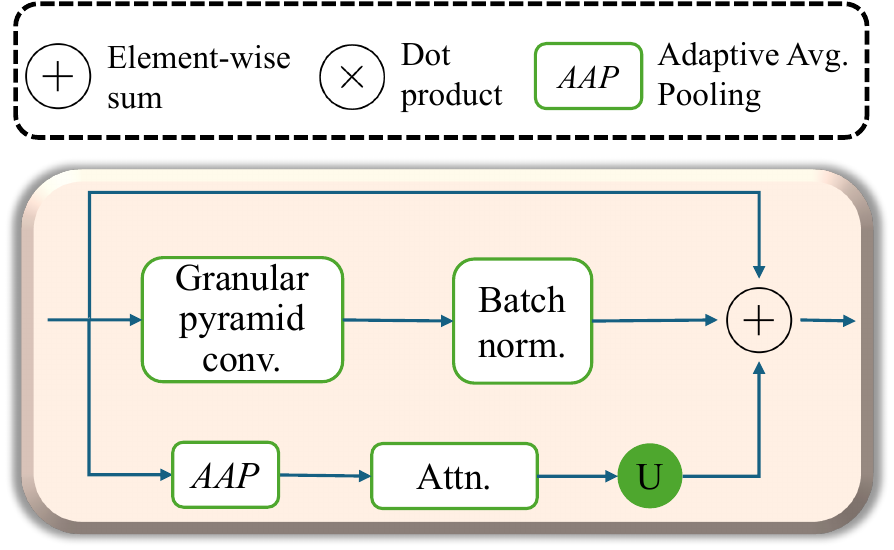} 
    \\
    {(a) Granular pyramid convolution with efficient attention.} % 
    \vspace{0.5cm}
    % \end{subfigure}

    \centering
    \includegraphics[width=0.9\linewidth]{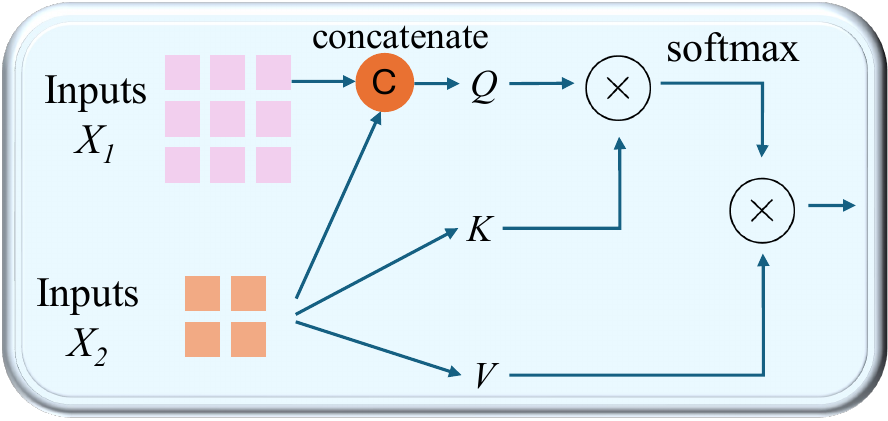}
    \\
    {(b) Cross-scale attention.} % Subfigure name
        
    \caption{\textbf{Illustration of GPC and CSA for multi-scale feature fusion.} For  cross-scale attention, $Q$ is computed with combined $X_1$ and $X_2$ while $K$, $V$ are computed with $X_2$ only.
    }
    \label{fig:GPC_CA}
\end{figure}%
}

\subsection{Multi-scale Feature Fusion} 
\label{sec:featurefusion}
Successful salient object detection necessitates simultaneous global localization \cite{wu2022edn} and multi-scale feature learning \cite{pang2020multi}.
Effectively extracting both, while maintaining efficiency, presents a significant challenge due to computational constraints.
In response, we have developed two distinct strategies: CNN-based (\secref{sec:gpc}) and transformer-based (\secref{sec:ca}) modules, designed specifically for low-level and high-level feature fusion, respectively.
Further details of these modules are discussed below.

\subsubsection{Granular pyramid convolution with efficient self-attention}
\label{sec:gpc}

% For low-level feature fusion,
%Inspired by CNN-based ASPP \cite{chen2017deeplab} for semantic  segmentation, 
We introduce an efficient {\PCASA} module for low-scale feature fusion as shown in \figref{fig:GPC_CA}(a).
It consists of a multi-scale feature extraction branch and an efficient global attention.
For the global attention module, we first apply adaptive average pooling to downsample the input to $m\times m$, thereby reducing computational overhead.
Attention is computed on the downsampled feature and then upsampled via bilinear interpolation.
The whole process can be elaborated as below:
%The downsampled feature is then used to compute the attention which is then upsampled to the same size as input features.
% \todo{PCASA and CSA needs a new name and an illustration}
\CheckRmv{
\begin{equation} \label{equ:lrsa}
\begin{aligned}
    F_{ds} &= {AAP}(F_{in}, (m \times m)) \\
    % Q, K, V &= F_{ds} (W^Q, W^K, W^V) \\
    % Att_P &= {\rm softmax}(\frac{QK^T}{\sqrt{d_k}})V \\
    Att_P &= F_{ds} + {\rm Attention}({\rm LayerNorm}(F_{ds})) \\
    F_{out}^{\rm A} &= {\rm Upsample}(Att_P)
\end{aligned}
\end{equation}
}
where $AAP$ is a 2D adaptive average pooling layer that pools the input feature to the size of $m\times m$.
$\rm Attention(\cdot)$ is the vanilla attention shown in \equref{equ:van_attn}.
$\rm Upsample(\cdot)$ upsamples attention to the same size as the input features $F_{in}$.
% $W^Q$, $W^K$, and $W^V$ are the weights.
% Note that $\rm LayerNorm$ is performed to $F_{ds}$ before computing attention $F_{Att}$. $\rm Linear(\cdot)$ is a linear transformation of the attention.

For the CNN block, the input features $F_{in}$ are first split into four feature maps along the channel dimension,
denoted as $F_1$, $F_2$, $F_3$ and $F_4$.
Unlike recent approaches \cite{wu2021mobilesal, wu2022edn} that evenly split channels we allocate ratios of 1/8, 1/8, 1/4, and 1/2 so that smaller dilation rates are applied to high-scale features.
%Unlike the recent approach \cite{wu2022edn} where input features are separated evenly, 
%we use [$\frac{1}{8}, \frac{1}{8}, \frac{1}{4}, \frac{1}{2}$] ratios to split them so that low dilation ratio is applied to high-scale features.
We concatenate the features of each split followed by a $1\times1$ convolution. The above processes are elaborated as below:
\CheckRmv{
\begin{equation}
\begin{aligned}
% C_1 &= {\rm Conv}_{3\times 3}^{a_1}(F_1), \\
C_i &= {\rm Conv}_{3\times 3}^{a_i}(F_i),\quad i\in \{1,2,3,4\} \\
F_{out}^{\rm C} &= {\rm Conv}_{1\times 1}(\rm Concat(C_1, C_2, C_3, C_4))
\end{aligned}
\end{equation}
}
where ${\rm Conv}_{3\times 3}^{a_i}(\cdot)$ is a $3\times 3$ atrous convolution with an atrous rate of $a_i$ followed by batch normalization.

Finally, we add a residual connection to aggregate the output feature $F_{out}$, which is computed as 
\CheckRmv{
\begin{equation}\label{equ:PCASA}
    F_{out} =  F_{out}^{\rm A} + F_{out}^{\rm C} + F_{in}
\end{equation}
}

\subsubsection{Cross-scale attention mechanism}
\label{sec:ca}
For high-level features, the spatial resolution is significantly reduced compared to the original image, which enables the deployment of attention mechanisms even with limited computational resources.
Consequently, we have developed a CSA block for high-level feature fusion, as illustrated in \figref{fig:GPC_CA}(b).
Unlike traditional attention mechanisms that first concatenate input features of different scales and then compute \(Q\), \(K\), and \(V\), our cross-level attention approach computes \(Q\) using combined input features, while \(K\) and \(V\) are derived solely from high-level features.
This approach is formulated as follows:

% \CheckRmv{
% \begin{equation}
% \begin{aligned}
%     Q_{CA} &= ({\rm Concat}(X_1, X_2)W^Q \\
%     K_{CA}, V_{CA} &= X_2(W^K, W^V)
% \end{aligned}
% \end{equation}
% }
\CheckRmv{
\begin{equation}
\begin{aligned}
    Q &= {\rm Concat}(X_1, X_2)W^Q \\
    (K, V) &= X_2(W^K, W^V)
\end{aligned}
\end{equation}
}
where $X_1$, $X_2$ are the flattened low-level and high-level features, respectively. $\rm LayerNorm(\cdot)$ is performed before calculating $Q$, $K$ and $V$.
% Attention and final output are calculated in the same way as regular transformers.
% $W^Q$, $W^K$, $W^V$ are the weights for the query, key and value, respectively. 

% Then, the output is computed as
% \CheckRmv{
% \begin{equation}
% Att_C = {\rm Linear}({\rm softmax})
% \end{equation}
% }

This cross-scale attention mechanism significantly reduces the computational burden.
In \equref{equ:dhl}, the scales of \(E_3\) and \(E_4\) are \(\frac{1}{16}\) and \(\frac{1}{32}\), respectively.
Consequently, the scale \(X_2\) constitutes only one fifth of \(\text{Concat}(X_1, X_2)\), reducing the complexity of the  cross-scale attention to just \(\frac{1}{25}\) of that observed in vanilla attention mechanisms.
Similarly to standard transformer blocks, attention is computed as outlined in \equref{equ:van_attn}.
Finally, an FFN comprising two linear layers with a residual connection is deployed to compute the fused features.

\subsubsection{Video feature fusion}

For video salient object detection, our framework adopts a two-stream architecture that processes both RGB frames and optical flow information to capture spatial-temporal dependencies. 
The fusion of RGB and optical flow features occurs at multiple hierarchical levels within our granularity-aware connections. 

At low-level stages ($E_1$ and $E_2$), we employ a simple yet effective fusion strategy that combines additive and multiplicative attention mechanisms before applying the granular pyramid convolution. 
Specifically, the optical flow features are first passed through a sigmoid activation to generate attention weights, which are then used to modulate the RGB features through element-wise multiplication. 
The final fused features combine both the attention-modulated RGB features and the original features from both modalities. 
This fusion mechanism allows the optical flow to serve as an attention gate that highlights motion-relevant regions while preserving complementary information from both streams.

At high-level stages ($E_3$ and $E_4$), we leverage the same cross-scale attention (CSA) modules used in our granularity-aware connections. 
The CSA mechanism naturally accommodates the fusion of multi-modal features by treating RGB and optical flow features as different input sequences. 
The CSA module computes cross-attention between RGB and flow features, enabling the model to capture long-range temporal dependencies and motion-guided spatial attention.

This hierarchical fusion strategy aligns with our granularity-aware paradigm: low-level fusion preserves fine-grained motion details essential for accurate boundary delineation, while high-level fusion captures coarse temporal semantics for robust object localization. The fused features are then processed through the same decoder structure as described in \secref{sec:decoder}, maintaining computational efficiency while enhancing temporal consistency in video salient object detection.

% The output feature is then computed as
% \CheckRmv{
% \begin{equation}
% \begin{aligned}
%     F_{Att}^{CA} &= {\rm softmax}(\frac{Q_{{CA}}K_{{CA}}^T}{\sqrt{d_k}})V_{{CA}} \\
%     F_out^CA &= F_out^CA
% \end{aligned}
% \end{equation}
% }
% where $X_1$, $X_2$ are the flattened low-level and high-level features, respectively. $W^Q$, $W^K$, $W^V$ are the weights for the query, key and value, respectively. 

\CheckRmv{%
\begin{figure}[!t]
    \centering
    \includegraphics[width=\linewidth]{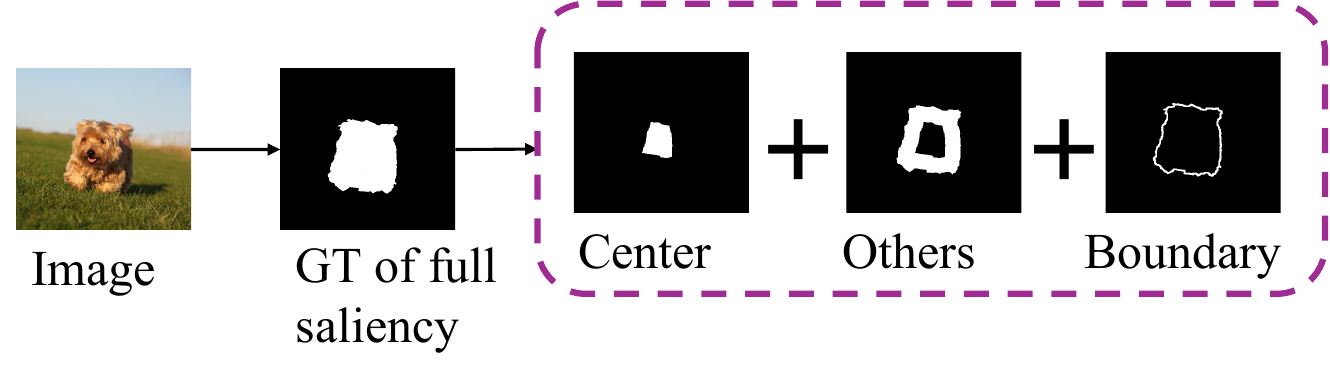}
    \caption{\textbf{Illustration of decomposing the foreground of full saliency map into multi-granularity regions: center, boundary and others}. 
    Black and white regions represent background and foreground, respectively.}
    \label{fig:decomposition}
    % \todo{change figure locations.}
\end{figure}%
}

\subsection{Granularity-aware Deep Supervision} \label{sec:framework}
Based on various encoder-decoder structures depicted in \figref{fig:ed-archs}, a deep supervision mechanism can be employed to leverage decoder side-outputs effectively.
Existing methods such as HED \cite{xie2015holistically}, U-Net \cite{wu2022edn}, and variants of U-Net like CPD \cite{wu2019cascaded}, typically utilize the full saliency map to supervise side-outputs at different scales.
Some recent methods  employ the edge supervision \cite{zhao2019egnet} in the low-level features or apply label decoupling strategy with an iterative training strategy. 
In contrast, our approach, as illustrated in \figref{fig:our-network}, proposes using distinct ground truths (center, edge, others, full) to supervise different outputs in a single stage, enhancing the specificity and effectiveness of the training process for lightweight SOD.

\subsubsection{Decomposition of ground-truth saliency map}
According to the Euclidean distance to the nearest background pixel, each pixel of the saliency foreground is classified into three regions \cite{wu2022edn}: the boundary, which is close to the background; the center, which is far from the background; and others, which are located in the middle of an object.
Specifically, the boundary region comprises pixels that are less than five pixels away from the closest background pixel. Pixels that rank in the top $20$\% in terms of distance from the nearest background pixel constitute the center region.
Any foreground pixels that do not qualify for inclusion in either the boundary or center regions are categorized into the other region.
The aggregation of the center and other regions is referred to as the boundary-others region.
The center region represents the abstract location of the object, while the boundary delineates the fine-grained edges of the object.
For illustration, an example is provided in \figref{fig:decomposition}.

Based on this classification, we employ low-granularity center saliency to supervise the high-level side-output, and high-granularity boundary and others saliency to supervise the low-level side-output. The final output is supervised using the full saliency map.

\subsubsection{Loss function}
The loss function combines the binary cross-entropy loss and Dice Loss \cite{milletari2016v}, defined as

% why alignment is center, different from other equations.
\CheckRmv{
\begin{equation}
\begin{aligned}
\mathcal{L}_{bce} &= -G\log P - (1 - G)\log (1-P) \\
\mathcal{L}_{dice} &= 1 - \frac{2 \cdot G\cdot P}{||G|| + ||P||} \\
\mathcal{L} &= \mathcal{L}_{bce} + \mathcal{L}_{dice}
\end{aligned}
\end{equation}
}
where $P$ and $G$ denote the predicted and ground-truth saliency map, respectively.
``$\cdot$'' operation is the dot product.
$||\cdot||$ denotes the $\ell_{1}$ norm.
$L_{bce}$, $L_{dice}$ and $L$ represent the binary cross-entropy loss, dice loss and combined loss, respectively. The Dice loss is an effective way to address class-imbalance datasets.

There are two side-outputs and one final output and the overall loss that we use for training is computed as
\begin{equation}
\mathcal{L}_{overall} = \sum_{i=1}^3 \mathcal{L}(P_i, G_i)
\end{equation}
where $G_1$, $G_2$ and $G_3$ are the ground-truth boundary-others saliency map, center saliency map and full saliency map, respectively.
$P_i$ is the corresponding predicted saliency map calculated from decoder side-outputs $D_1$, $D_2$ and $D_3$ in \equref{equ:d12} and \equref{equ:d3}, shown as
\begin{equation} \label{equ:deep_supvision}
P_i = \sigma({\rm Upsample}({\rm Conv}_{1\times 1}(D_i))), \quad i\in \{1,2,3\}
\end{equation}
where ${\rm Conv}_{1\times 1}(\cdot)$ denotes a convolutional layer without normalization and activation.
$\rm Upsample(\cdot)$ upsamples input features to the same resolution as the full saliency map using bilinear interpolation.
$\sigma(\cdot)$ is the standard sigmoid function.
% $D_1$ and $D_2$ are the low-level and high-level side-outputs, respectively. $D_3$ is the final output of the decoder.

\section{Experiments}

\subsection{Experimental Setup}

% Table generated by Excel2LaTeX from sheet '\sArt Comparison'
\begin{table*}[t]
  \centering
  \setlength\tabcolsep{1pt}
  \caption{\textbf{Comparison of GAPNet with \sArt heavyweight and lightweight SOD methods}.  
  The best performance in each row among lightweight models is highlighted in bold.}
  \resizebox{\textwidth}{!}{%
    \begin{tabular}{c|c|cccccccccc|cccccccc}
    \Xhline{1pt}
    \multicolumn{2}{c|}{\multirow{2}[2]{*}{Method}} &
      \multicolumn{10}{c|}{Heavyweight models (\# Param $>$ 20 M)} &
      \multicolumn{8}{c}{Lightweight models (\# Param $<$ 2 M)}\bigstrut[b]\\
\cline{3-20}    \multicolumn{2}{c|}{} &
      CPD & PoolNet & ITSD & MINet & VST & CTD & ICON & EDN & SRF & PiNet &
      HVPNet & CSNet & SAMNet & EDN\text{-}Lite & ELWNet & LARNet & ADMNet+ & Ours\bigstrut[t]\\
    \multicolumn{2}{c|}{\# Reference} &
      \cite{wu2019cascaded} & \cite{liu2019simple} & \cite{zhou2020interactive} &
      \cite{pang2020multi} & \cite{liu2021visual} & \cite{zhao2021complementary} &
      \cite{zhuge2022salient} & \cite{wu2022edn} & \cite{yun2023towards} & \cite{wang2024feature} &
      \cite{liu2020lightweight} & \cite{cheng2021highly} & \cite{liu2021samnet} &
      \cite{wu2022edn} & \cite{wang2023elwnet} & \cite{wang2023larnet} & \cite{zhou2024admnet} & -\bigstrut[b]\\
    \Xhline{1pt}
    \multicolumn{2}{c|}{\# Param (M)} &
      47.85 & 68.26 & 26.47 & 162.38 & 44.56 & 24.64 & 33.09 & 42.85 & 91.59 & 27.20 &
      1.24 & 0.14 & 1.33 & 1.80 & 0.07 & 0.09 & 0.84 & 1.99\\
    \multicolumn{2}{c|}{FLOPs (G)} &
      17.74 & 89.06 & 15.95 & 87.02 & 41.36 & 12.35 & 20.91 & 20.37 & 21.77 & -- &
      1.13 & 0.57 & 0.54 & 1.02 & 0.38 & 0.82 & 3.30 & 1.26\\
    \multicolumn{2}{c|}{Speed (FPS)} &
      60 & 83 & 77 & 62 & 73 & 148 & 89 & 77 & 39 & -- &
      749 & 675 & 459 & 652 & -- & -- & 115 & 571\\
    \hline
    \multirow{6}[1]{*}{DUTS-TE} & $F_\beta^{\text{max}}$ &
      0.865 & 0.874 & 0.882 & 0.880 & 0.890 & 0.896 & 0.891 & 0.893 & 0.913 & 0.865 &
      0.837 & 0.804 & 0.833 & 0.856 & -- & -- & 0.840 & \textbf{0.867}\\
          & $F_\beta^{w}$ &
      0.794 & 0.806 & 0.822 & 0.824 & 0.827 & 0.846 & 0.835 & 0.844 & 0.871 & 0.817 &
      0.730 & 0.643 & 0.729 & 0.789 & -- & -- & 0.767 & \textbf{0.804}\\
          & MAE &
      0.043 & 0.040 & 0.041 & 0.038 & 0.038 & 0.034 & 0.037 & 0.035 & 0.027 & 0.041 &
      0.058 & 0.075 & 0.058 & 0.045 & 0.075 & 0.069 & 0.052 & \textbf{0.042}\\
          & $S_\alpha$ &
      0.869 & 0.883 & 0.884 & 0.883 & 0.896 & 0.892 & 0.888 & 0.892 & 0.910 & 0.864 &
      0.849 & 0.822 & 0.849 & 0.862 & -- & 0.820 & 0.849 & \textbf{0.872}\\
          & $E_\xi^{\text{max}}$ &
      0.914 & 0.923 & 0.930 & 0.927 & 0.939 & 0.935 & 0.932 & 0.934 & 0.952 & 0.911 &
      0.899 & 0.875 & 0.902 & 0.910 & -- & -- & 0.892 & \textbf{0.922}\\
          & $E_\xi^{\text{mean}}$ &
      0.898 & 0.904 & 0.914 & 0.917 & 0.919 & 0.929 & 0.923 & 0.925 & 0.943 & 0.906 &
      0.860 & 0.820 & 0.860 & 0.895 & -- & -- & 0.882 & \textbf{0.910}\bigstrut[b]\\
    \hline
    \multirow{6}[2]{*}{DUT-OMRON} & $F_\beta^{\text{max}}$ &
      0.797 & 0.792 & 0.818 & 0.795 & 0.822 & 0.818 & 0.821 & 0.821 & 0.825 & 0.793 &
      0.796 & 0.761 & 0.795 & 0.783 & -- & -- & 0.797 & \textbf{0.806}\\
          & $F_\beta^{w}$ &
      0.719 & 0.729 & 0.750 & 0.738 & 0.755 & 0.762 & 0.761 & 0.770 & 0.784 & -- &
      0.700 & 0.620 & 0.699 & 0.721 & -- & -- & 0.729 & \textbf{0.738}\\
          & MAE &
      0.056 & 0.055 & 0.061 & 0.056 & 0.058 & 0.052 & 0.057 & 0.050 & 0.043 & 0.055 &
      0.064 & 0.080 & 0.065 & 0.058 & 0.083 & 0.080 & 0.058 & \textbf{0.057}\\
          & $S_\alpha$ &
      0.825 & 0.836 & 0.840 & 0.833 & 0.850 & 0.844 & 0.844 & 0.849 & 0.861 & 0.821 &
      0.831 & 0.805 & 0.830 & 0.824 & -- & 0.797 & 0.826 & \textbf{0.833}\\
          & $E_\xi^{\text{max}}$ &
      0.868 & 0.871 & 0.880 & 0.869 & 0.888 & 0.881 & 0.884 & 0.885 & 0.894 & 0.863 &
      0.876 & 0.853 & 0.877 & 0.860 & -- & -- & 0.869 & \textbf{0.876}\\
          & $E_\xi^{\text{mean}}$ &
      0.847 & 0.854 & 0.865 & 0.860 & 0.871 & 0.875 & 0.876 & 0.878 & 0.884 & 0.859 &
      0.839 & 0.801 & 0.841 & 0.848 & -- & -- & 0.857 & \textbf{0.866}\bigstrut[b]\\
    \hline
    \multirow{6}[1]{*}{HKU-IS} & $F_\beta^{\text{max}}$ &
      0.925 & 0.930 & 0.934 & 0.934 & 0.942 & 0.940 & 0.939 & 0.940 & 0.947 & 0.928 &
      0.914 & 0.896 & 0.914 & 0.922 & -- & -- & 0.918 & \textbf{0.929}\\
          & $F_\beta^{w}$ &
      0.875 & 0.881 & 0.894 & 0.897 & 0.897 & 0.909 & 0.902 & 0.908 & 0.915 & 0.896 &
      0.840 & 0.777 & 0.837 & 0.877 & -- & -- & 0.872 & \textbf{0.889}\\
          & MAE &
      0.034 & 0.033 & 0.031 & 0.029 & 0.030 & 0.027 & 0.029 & 0.027 & 0.024 & 0.030 &
      0.045 & 0.060 & 0.045 & 0.035 & 0.051 & 0.046 & 0.036 & \textbf{0.032}\\
          & $S_\alpha$ &
      0.905 & 0.915 & 0.917 & 0.919 & 0.928 & 0.921 & 0.920 & 0.924 & 0.931 & 0.904 &
      0.899 & 0.881 & 0.898 & 0.906 & -- & 0.883 & 0.901 & \textbf{0.914}\\
          & $E_\xi^{\text{max}}$ &
      0.950 & 0.954 & 0.960 & 0.960 & 0.968 & 0.961 & 0.960 & 0.962 & 0.969 & 0.951 &
      0.946 & 0.933 & 0.946 & 0.948 & -- & -- & 0.946 & \textbf{0.957}\\
          & $E_\xi^{\text{mean}}$ &
      0.938 & 0.939 & 0.947 & 0.952 & 0.952 & 0.956 & 0.953 & 0.955 & 0.960 & 0.946 &
      0.914 & 0.883 & 0.912 & 0.936 & -- & -- & 0.934 & \textbf{0.947}\\
    \hline
    \multirow{6}[1]{*}{ECSSD} & $F_\beta^{\text{max}}$ &
      0.939 & 0.943 & 0.947 & 0.946 & 0.951 & 0.949 & 0.950 & 0.950 & 0.957 & 0.935 &
      0.927 & 0.912 & 0.926 & 0.934 & -- & -- & 0.922 & \textbf{0.938}\\
          & $F_\beta^{w}$ &
      0.898 & 0.896 & 0.910 & 0.911 & 0.910 & 0.915 & 0.918 & 0.918 & 0.926 & 0.902 &
      0.854 & 0.806 & 0.858 & 0.890 & -- & -- & 0.871 & \textbf{0.898}\\
          & MAE &
      0.037 & 0.039 & 0.035 & 0.034 & 0.034 & 0.032 & 0.032 & 0.033 & 0.027 & 0.039 &
      0.053 & 0.066 & 0.051 & 0.043 & 0.061 & 0.055 & 0.051 & \textbf{0.040}\\
          & $S_\alpha$ &
      0.918 & 0.921 & 0.925 & 0.925 & 0.932 & 0.925 & 0.929 & 0.927 & 0.936 & 0.910 &
      0.903 & 0.893 & 0.907 & 0.911 & -- & 0.888 & 0.900 & \textbf{0.916}\\
          & $E_\xi^{\text{max}}$ &
      0.951 & 0.952 & 0.959 & 0.957 & 0.964 & 0.956 & 0.960 & 0.958 & 0.965 & 0.948 &
      0.940 & 0.931 & 0.944 & 0.944 & -- & -- & 0.933 & \textbf{0.950}\\
          & $E_\xi^{\text{mean}}$ &
      0.942 & 0.940 & 0.947 & 0.950 & 0.951 & 0.950 & 0.954 & 0.951 & 0.957 & 0.944 &
      0.911 & 0.886 & 0.916 & 0.933 & -- & -- & 0.914 & \textbf{0.941}\bigstrut[b]\\
    \hline
    \multirow{6}[2]{*}{PASCAL-S} & $F_\beta^{\text{max}}$ &
      0.859 & 0.862 & 0.870 & 0.865 & 0.875 & 0.877 & 0.876 & 0.879 & 0.892 & 0.858 &
      0.838 & 0.826 & 0.836 & 0.852 & -- & -- & 0.827 & \textbf{0.860}\\
          & $F_\beta^{w}$ &
      0.794 & 0.793 & 0.812 & 0.809 & 0.816 & 0.822 & 0.818 & 0.827 & 0.848 & 0.807 &
      0.746 & 0.691 & 0.738 & 0.788 & -- & -- & 0.752 & \textbf{0.793}\\
          & MAE &
      0.071 & 0.075 & 0.066 & 0.064 & 0.062 & 0.061 & 0.064 & 0.062 & 0.051 & 0.069 &
      0.090 & 0.104 & 0.092 & 0.073 & 0.102 & 0.096 & 0.088 & \textbf{0.073}\\
          & $S_\alpha$ &
      0.848 & 0.849 & 0.859 & 0.856 & 0.872 & 0.863 & 0.861 & 0.865 & 0.881 & 0.837 &
      0.830 & 0.814 & 0.826 & 0.842 & -- & 0.810 & 0.815 & \textbf{0.843}\\
          & $E_\xi^{\text{max}}$ &
      0.891 & 0.891 & 0.908 & 0.903 & 0.918 & 0.906 & 0.908 & 0.908 & 0.928 & 0.889 &
      0.872 & 0.860 & 0.870 & 0.890 & -- & -- & 0.862 & \textbf{0.890}\\
          & $E_\xi^{\text{mean}}$ &
      0.882 & 0.880 & 0.895 & 0.896 & 0.902 & 0.901 & 0.899 & 0.902 & 0.919 & 0.886 &
      0.844 & 0.815 & 0.839 & 0.878 & -- & -- & 0.851 & \textbf{0.881}\bigstrut[b]\\
    \Xhline{1pt}
    \end{tabular}%
  }
  \label{tab:sota}%
\end{table*}

\para{Implementation details.}
The proposed model is implemented in PyTorch \cite{paszke2019pytorch} with a single NVIDIA RTX3090 GPU.
Training is carried out over $30$ epochs using the Adam optimizer \cite{kingma2015adam}, with parameters set to $\beta_1=0.9$, $\beta_2=0.99$, a weight decay of $10^{-4}$, and a batch size of $32$.
We employ a polynomial learning rate scheduler with an initial learning rate  of $1.7 \times 10^{-4}$ and a power of $0.9$.
The adaptive pooling size of the \(\text{\PCASA}\) module is set to \(m=7\) \equref{equ:lrsa}.
During training, the input images are resized to $320\times320$, $352\times352$, and $384\times384$ for augmentation purposes.
During inference, images are resized to $384\times384$.
The CSA and GPC modules are highly efficient, with just 0.065M and 0.020M parameters.
For video SOD, we first train our model using static images of DUTS training set and then finetune on the video dataset. Following previous popular works \cite{ji2021full, fu2021siamese, liu2024learning}, we apply FlowNet 2.0 \cite{ilg2017flownet} to generate the offline optical flows.
The video SOD training hyper-parameters match those of image SOD, except that the learning rate is reduced by a factor of ten.

\para{SOD Datasets.}
The proposed method has been tested on five commonly-used datasets, including three large datasets: DUTS \cite{wang2017learning}, DUT-OMRON \cite{yang2013saliency}, HKU-IS \cite{li2015visual}, and two smaller datasets: ECSSD \cite{yan2013hierarchical} and PASCAL-S \cite{li2014secrets}.
These datasets comprise $15572$, $5168$, $4447$, $1000$, and $850$ natural images with corresponding pixel-level labels, respectively.
Following methodologies from prior studies \cite{wang2018detect,zeng2018learning,wang2023pixels}, we train our model on the DUTS training set, which contains $10553$ images, and evaluate it on the DUTS test set (DUTS-TE, $5019$ images) and the other four datasets. 

\para{Video SOD Datasets.}
We utilize four commonly-used datasets DAVSOD \cite{fan2019shifting}, DAVIS \cite{perazzi2016benchmark}, SegTrack-V2 \cite{li2013video}, and ViSal \cite{wang2015consistent} to construct the experiments. 
Following other approaches, our model is also trained on the training set of DAVSOD \cite{fan2019shifting} and DAVIS \cite{perazzi2016benchmark}, which have 91 clips in total. Other data are for testing. For DAVSOD, we use the easy set of 35 clips for testing. 

\para{Evaluation Criteria.}
We employ six widely-used metrics to evaluate all methods, which include
the maximum F-measure score ($F_\beta^{\text{max}}$),
weighted F-measure score ($F_\beta^w$) \cite{margolin2014evaluate},
mean absolute error (MAE),
S-measure ($S_\alpha$) \cite{fan2017structure},
maximum E-measure ($E_\xi^{\text{max}}$), 
and mean E-measure ($E_\xi^{\text{mean}}$) \cite{fan2018enhanced}.
Except for MAE, a higher value indicates better performance for all metrics.
${\rm F}$-measure is the weighted harmonic mean of precision and recall and can be calculated as
\begin{equation}
F_{\beta} = \frac{(1 + \beta^2) \times {\rm Precision} \times {\rm Recall}}{\beta^2 \times {\rm Precision} + {\rm Recall}}
\end{equation}
where $\beta^2 = 0.3$ to emphasize the importance of precision, following previous studies \cite{hou2019deeply,liu2019simple,liu2020picanet,zhang2017amulet}.
$F_\beta^{\text{max}}$ is the maximum $F_\beta$ under different binary thresholds. $F_\beta^w$ solves the problems of ${\rm F}$-measure that may cause three types of flaw, \ie, interpolation, dependency, and equal-importance \cite{margolin2014evaluate}.

$\rm MAE$ measures the similarity between the predicted saliency map $P$ and the ground-truth saliency map $G$, which can be computed as
\begin{equation}
{\rm MAE}(P, G) = \frac{1}{HW} \sum_{i=1}^{H} \sum_{j=1}^{W} \| P_{i,j} - G_{i,j} \|
\end{equation}
where $H$ and $W$ denote the height and width of the saliency map, respectively.

${\rm S}$-measure ($S_\alpha$) \cite{fan2017structure} and ${\rm E}$-measure ($E_\xi$) \cite{fan2018enhanced} 
have been increasingly popular for SOD evaluation recently \cite{pang2020multi,zhao2020depth, wang2023larnet}.
${\rm S}$-measure calculates the structural similarity between the predicted saliency map and the ground-truth map.
${\rm E}$-measure computes the similarity for the predicted map binarized by different thresholds and the binary ground-truth map.
Thus, they are significant alternatives that could provide more comprehensive SOD evaluations.
In this paper, we compute the maximum and average E-measures ($E_\xi^{\text{max}}$, $E_\xi^{\text{mean}}$) among all binary thresholds.
We use the official codes from \cite{fan2017structure, fan2018enhanced} to compute the above metrics.
%$F_\beta^w$, $S_\alpha$, $E_\xi^{\text{max}}$ and $E_\xi^{\text{mean}}$.

% Comparison of OUR MODEL with \sArt heavyweight and lightweight SOD methods. The best performance in each row among lightweight models is highlighted in bold. Speed row contains FPS with batch size 1 and 30 ("-" indicates that GPU is out of memory). Input image size for model efficiency testing is $224\times224$ for CSNet, SRF and VST; $288\times288$ for ITSD-R; $320\times320$ for EDN-Lite, EDN-V, LARNet, MINet-R, ELWNet, and Ours; $336\times336$ for SAMNet and HVPNet; $352\times352$ for CPD-R, LDF and ICON-R.

\begin{figure*}[!t]
    \centering
    \footnotesize
    \renewcommand{\arraystretch}{0.3}
    \setlength{\tabcolsep}{0.2mm}
    % \resizebox{\textwidth}{!}{%
    \begin{tabular}{cccc}
        \includegraphics[height=0.23\textwidth]{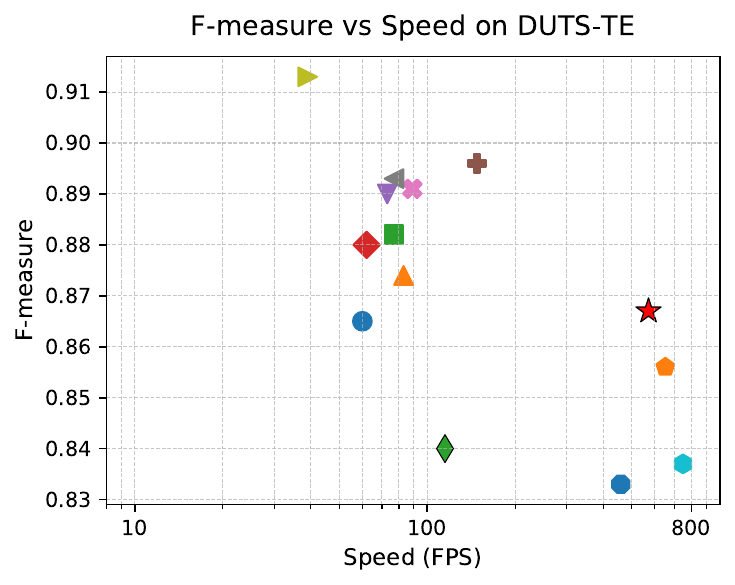} &
        \includegraphics[height=0.23\textwidth]{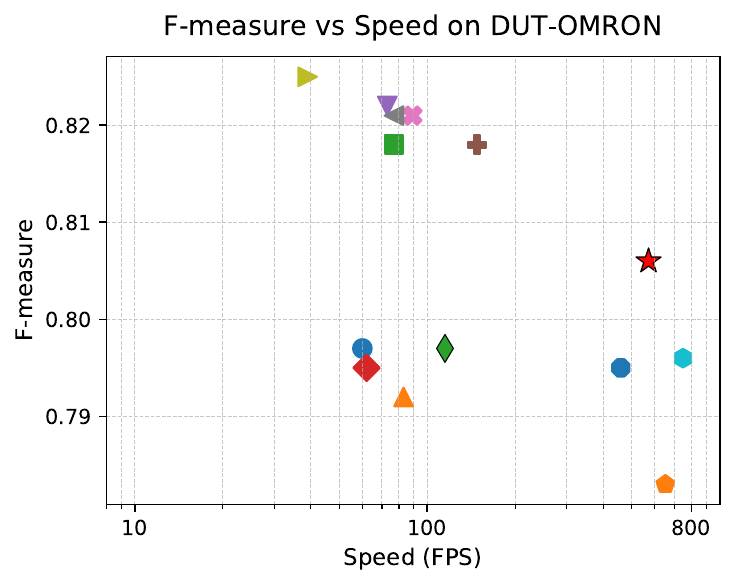} &
        \includegraphics[height=0.23\textwidth]{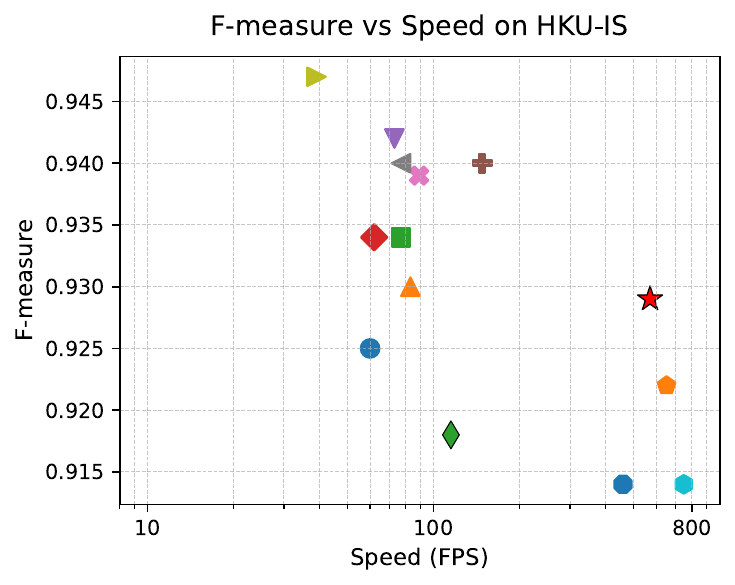} &
        \includegraphics[height=0.23\textwidth]{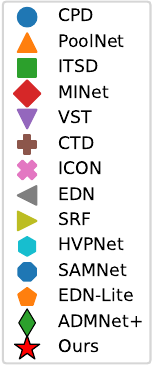}
        \\
        (a) DUTS-TE \cite{wang2017learning} &
        (b) DUT-OMRON \cite{yang2013saliency} &
        (c) HKU-IS \cite{li2015visual}
    \end{tabular}
    % }
    \caption{\textbf{Speed and accuracy comparison with \sArt SOD methods}. 
    Our model outperforms all the lightweight models and some of the heavyweight models. Inference speed is plotted using logarithm with base $\rm 10$.
    }
    \label{fig:AVP}
\end{figure*}

\subsection{Experimental Comparisons} \label{sec:comparison}

\para{Image SOD.}
We compare our model against nine heavyweight models with over 10M parameters and six lightweight models with no more than 10M parameters.
For competing models that offer both ResNet-50 and VGG-16 backbones, the ResNet-50 backbone is utilized.
For lightweight models, all models are with the MobileNetV2 backbone, except that CSNet, ELWNet, and LARNet designed their backbones for extremely lightweight SOD.
For a fair comparison, we use the saliency maps provided by the official repositories of the benchmarking methods and use the same code for evaluation.
To assess model efficiency, we re-implement these models on the same workstation equipped with a single NVIDIA RTX3090 GPU.
%Methods for calculating FLOPs and inference speed are derived from \href{https://github.com/facebookresearch/fvcore}{fvcore} and \href{https://github.com/yuhuan-wu/MobileSal}{MobileSal}, respectively.
The input image sizes for the competing models adhere to the default settings specified in their original publications.

For LARNet \cite{wang2023larnet} and ELWNet \cite{wang2023elwnet}, where no official codes or saliency maps are available, we directly extract data on the number of parameters, FLOPs, and selected performance metrics from the published papers.

\para{Video SOD.}
We compare our model against several recent heavyweight models. 
For fair comparison, we re-implement two recent strongest heavyweight models \cite{fu2021siamese, zhao2024motion} in the lightweight setting, \ie, replacing the backbone and 3$\times$3 convolutions with MobileNetV2 and 3$\times$3 depth-wise convolutions, respectively.
Following previous popular works \cite{fan2019shifting}, we apply S-measure, maximum F-measure, and MAE as the evaluation metrics.

%---------------------------------------------

\begin{table*}[t]
  \centering
  \small                      
  \caption{\textbf{Comparison with state-of-the-art methods on video SOD.} The best performance of lightweight models is marked in bold.}
  \setlength\tabcolsep{4pt}   
  \renewcommand{\arraystretch}{1.15}

  \resizebox{\textwidth}{!}{
  \begin{tabular}{lcc|ccc|ccc|ccc|ccc}
    \toprule
    \multirow{2}{*}{Method} & \multirow{2}{*}{Param (M)} & \multirow{2}{*}{FPS} &
    \multicolumn{3}{c|}{DAVIS \cite{perazzi2016benchmark} } &
    \multicolumn{3}{c|}{ViSal \cite{wang2015consistent}} &
    \multicolumn{3}{c|}{DAVSOD \cite{fan2019shifting}} &
    \multicolumn{3}{c}{SegTrack-V2 \cite{li2013video}} \\
    \cline{4-6}\cline{7-9}\cline{10-12}\cline{13-15}
     &  &  & $S_{\alpha}$ & $F_{\beta}^{max}$ & MAE &
                   $S_{\alpha}$ & $F_{\beta}^{max}$ & MAE &
                   $S_{\alpha}$ & $F_{\beta}^{max}$ & MAE &
                   $S_{\alpha}$ & $F_{\beta}^{max}$ & MAE \\
    \midrule
    \multicolumn{15}{c}{\textbf{Heavyweight models}}\\
    \midrule
    MGA~\cite{li2019motion}      & 87.5 & 14  & 0.910 & 0.892 & 0.022 & 0.940 & 0.936 & 0.017 & 0.741 & 0.643 & 0.083 & 0.880 & 0.829 & 0.027 \\
    RCR~\cite{yan2019semi}       & 51.5 & 27  & 0.886 & 0.848 & 0.027 & 0.922 & 0.906 & 0.027 & 0.741 & 0.653 & 0.087 & 0.843 & 0.782 & 0.035 \\
    SSAV~\cite{fan2019shifting}  & 59.0 & 20  & 0.893 & 0.861 & 0.028 & 0.943 & 0.939 & 0.020 & 0.724 & 0.603 & 0.092 & 0.851 & 0.801 & 0.023 \\
    LTSD~\cite{wang2020learning} &  --  & --  & 0.897 & 0.891 & 0.021 &  --  &  --  &  --  & 0.768 & 0.689 & 0.075 & 0.880 & 0.866 & 0.018 \\
    TENet~\cite{ren2020tenet}    &  --  & --  & 0.905 & 0.894 & 0.021 & 0.943 & 0.947 & 0.021 & 0.753 & 0.648 & 0.078 &  --  &  --  &  --  \\
    WSV~\cite{zhao2021weakly}    & 33.0 & 37  & 0.828 & 0.779 & 0.037 & 0.857 & 0.831 & 0.041 & 0.705 & 0.605 & 0.103 & 0.804 & 0.738 & 0.033 \\
    STVS~\cite{chen2021exploring}& 46.0 & 50  & 0.892 & 0.865 & 0.023 & 0.954 & 0.953 & 0.013 & 0.744 & 0.650 & 0.086 & 0.891 & 0.860 & 0.017 \\
    DCFNet~\cite{zhang2021dynamic}& 68.5& 28  & 0.914 & 0.900 & 0.016 & 0.952 & 0.953 & 0.010 & 0.741 & 0.660 & 0.074 & 0.883 & 0.839 & 0.015 \\
    FSNet~\cite{ji2021full}      & 97.9 & 28  & 0.920 & 0.907 & 0.020 &  --  &  --  &  --  & 0.773 & 0.685 & 0.072 &  --  &  --  &  --  \\
    CoSTFormer \cite{liu2024learning}      &  --  & 13  & 0.921 & 0.903 & 0.014 &  --  &  --  &  --  & 0.806 & 0.731 & 0.061 & 0.888 & 0.833 & 0.015 \\
    DMPN~\cite{chen2023dynamic}  &152.2 & 9   & 0.905 & 0.888 & 0.021 & 0.929 &  --  & 0.016 & 0.755 & 0.655 & 0.069 &  --  &  --  &  --  \\
    EESTI~\cite{chen2021exploring}& 46.2& 100 & 0.892 & 0.865 & 0.023 & 0.952 & 0.952 & 0.013 & 0.746 & 0.651 & 0.086 & 0.891 & 0.860 & 0.017 \\
    Li \textit{et~al.}~\cite{li2024novel} & -- & -- & 0.906 & 0.888 & 0.018 &  --  &  --  &  --  & 0.777 & 0.716 & 0.072 &  --  &  --  &  --  \\
    MMN~\cite{zhao2024motion}    & 49.0 & 69  & 0.897 & 0.877 & 0.020 & 0.947 & 0.948 & 0.012 & 0.777 & 0.708 & 0.065 & 0.886 & 0.850 & 0.014 \\
    \midrule
    \multicolumn{15}{c}{\textbf{Lightweight models}}\\
    \midrule
    JL-DCF-Light~\cite{fu2021siamese}  & 2.1 & 58  & 0.892 & 0.863 & 0.025 & 0.882 & 0.858 & 0.038 & \textbf{0.728} & \textbf{0.630} & \textbf{0.088} & 0.825 & 0.743 & 0.030 \\
    MMN-Light~\cite{zhao2024motion} & 3.1 & 340 & 0.861 & 0.822 & 0.025 & 0.884 & 0.864 & 0.035 & 0.700 & 0.593 & 0.089 & 0.843 & 0.786 & 0.023 \\
    \textbf{Ours}                & 3.8 & 349 & \textbf{0.893} & \textbf{0.864} & \textbf{0.021} & \textbf{0.886} & \textbf{0.867} & \textbf{0.033} & 0.706 & 0.597 & 0.089 & \textbf{0.862} & \textbf{0.804} & \textbf{0.021} \\
    \bottomrule
  \end{tabular}
   }   % 
  \label{tab:vsod_benchmark}
\end{table*}

\subsubsection{Quantitative comparison}
% On PASCAL-S dataset, the proposed model osutperforms \sArt using $F_\beta^{\text{max}}$ but outperforms with other five metrics.
% On SOC6K with complex scenarios, our model outperforms \sArt by 1.3\%, 1.3\% and 0.8\%, respectively.

\para{Image SOD.}
A comprehensive quantitative comparison of our model with competing methods is presented in \tabref{tab:sota}.
Our model consistently outperforms or matches other lightweight methods across the five datasets using all six metrics.
Specifically, our model surpasses the \sArt lightweight model EDN-Lite \cite{wu2022edn} by margins of 1.1\%, 2.3\%, 0.7\%, 0.4\%, and 0.8\% in terms of $F_\beta^{\text{max}}$ across the datasets.
Moreover, using $E_\xi^{\text{mean}}$, our model achieves performance improvements of 1.5\%, 1.8\%, 1.1\%, 0.8\%, and 0.3\%.
For $S_\alpha$, our model beats \sArt by 1.0\%, 0.9\%, and 0.8\% on the three large datasets, namely DUTS-TE, DUT-OMRON, and HKU-IS.
% In terms of $\rm MAE$ and $E_\xi^{\text{max}}$, our model matches \sArt performance on the PASCAL-S dataset and outperforms on the other four datasets. 
Notably, the most significant improvements are observed on the DUT-OMRON dataset, where $S_\alpha$, $E_\xi^{\text{max}}$, $E_\xi^{\text{mean}}$, $F_\beta^{\text{max}}$, $F_\beta^w$, and $\rm MAE$ are improved by 0.9\%, 1.6\%, 1.8\%, 2.3\%, 1.7\%, and 0.1\%, respectively.

In terms of model efficiency, our model possesses more parameters and is comparatively less efficient against extremely lightweight models CSNet \cite{cheng2021highly}, LARNet \cite{wang2023larnet}, and ELWNet \cite{wang2023elwnet}.
However, there is a notable accuracy gap between these models and ours.
For instance, the $\rm MAE$ of LARNet \cite{wang2023larnet} on DUTS-TE is 0.069, whereas our model achieves an $\rm MAE$ of 0.042.
Compared to lightweight models with similar parameters, including SAMNet \cite{liu2021samnet}, HVPNet \cite{liu2020lightweight}, and EDN-Lite \cite{wu2022edn}, our model exhibits slightly higher FLOPs and reduced inference speed.
This increased computational demand is attributed to the attention modules integrated into our model. In summary, our model establishes new benchmarks in \sArt performance for lightweight SOD models across all test cases, albeit at the expense of marginally higher computational overhead and slower inference speeds.

A comparison of the accuracy ($F_\beta^{\text{max}}$) and inference speed across three large datasets (e.g., DUTS-TE, DUT-OMRON, and HKU-IS) is depicted in \figref{fig:AVP}. It is evident that our model consistently surpasses other lightweight models across all datasets with significant improvements.
In certain test cases, our model achieves performance comparable to or even surpassing some heavyweight models, which exhibit considerably slower inference speeds.
For example, on the DUT-OMRON dataset, our model outperforms heavyweight models such as CPD \cite{wu2019cascaded}, PoolNet \cite{liu2019simple}, and MINet \cite{pang2020multi}.

\begin{figure}[!t]
    \centering
    \footnotesize
    \renewcommand{\arraystretch}{0.3}
    \setlength{\tabcolsep}{0.2mm}
    % \resizebox{\textwidth}{!}{%
    \begin{tabular}{ccc}
        \includegraphics[height=0.22\textwidth]{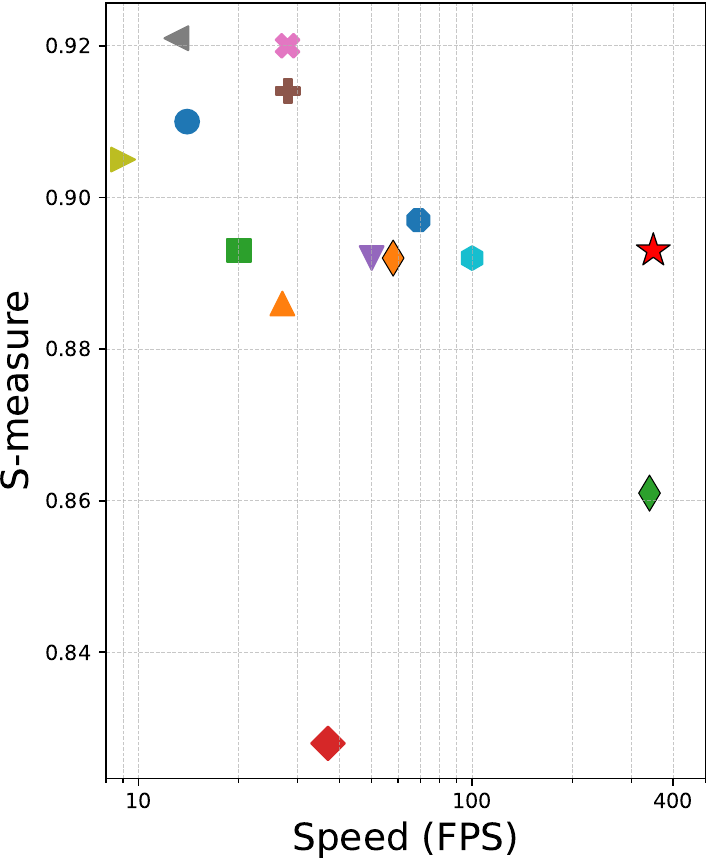} &
        \includegraphics[height=0.22\textwidth]{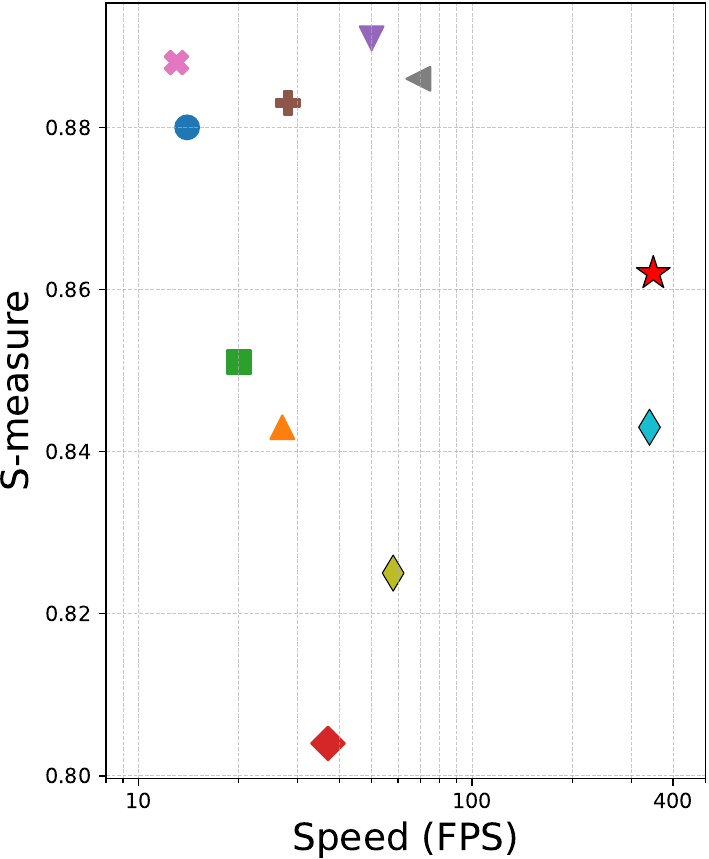} &
        \includegraphics[height=0.22\textwidth]{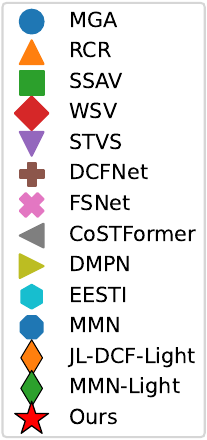}
        \\
        (a) DAVIS \cite{wang2017learning} &
        (b) SegTrack-V2 \cite{yang2013saliency} &
        %(c) HKU-IS \cite{li2015visual}
    \end{tabular}
    % }
    \caption{\textbf{Speed and accuracy comparison with \sArt video SOD methods}. 
    %Our model outperforms all the lightweight models and some of the heavyweight models. Inference speed is plotted using logarithm with base $\rm 10$.
    }
    \label{fig:AVP_video}
\end{figure}

\para{Video SOD.}
Results are shown in \tabref{tab:vsod_benchmark}.
We list a comparison of popular heavyweight and lightweight models in recent years.
From the results, we can find that our model outperforms the recent lightweight versions of JL-DCF \cite{fu2021siamese} and MMN \cite{zhao2024motion}. 
Although the lightweight JL-DCF achieves better performance than our model on the DAVSOD dataset, it falls short on other datasets, particularly on SegTrack-V2. 
Furthermore, our GAPNet operates significantly faster than the lightweight JL-DCF, highlighting its suitability for lightweight applications, especially on edge devices.
Compared to heavyweight models, our GAPNet demonstrates competitive performance while offering substantial efficiency advantages. 
Despite having significantly fewer parameters than heavyweight counterparts, 
our method operates at much higher inference speeds and achieves comparable or superior accuracy on most datasets especially DAVIS and SegTrack-V2. 
%GAPNet achieves comparable performance despite having significantly fewer parameters than heavyweight counterparts, 
This demonstrates that our granularity-aware paradigm effectively bridges the performance gap between lightweight and heavyweight approaches, making it highly suitable for real-time applications and resource-constrained environments without sacrificing detection quality.

Following the SOD part, we also illustrate the speed-accuracy comparison as shown in \figref{fig:AVP_video}.
Our method consistently occupies the upper-right corner of the accuracy-speed plots on both DAVIS and SegTrack-V2, delivering S-measure scores that rival or surpass heavyweight competitors while running an order of magnitude faster (300 FPS).
This clear dominance in the speed–accuracy Pareto front highlights the superior efficiency of the proposed framework over all lightweight baselines and many heavyweight models alike.

\newcommand{\AddImg}[1]{%
    \includegraphics[height=0.065\textwidth]{Imgs/examples/ILSVRC2012_test_00000354#1} &%
    \includegraphics[height=0.065\textwidth]{Imgs/examples/3059#1} &%
    \includegraphics[height=0.065\textwidth]{Imgs/examples/sun_bqsstriisebdjmnl#1} &%
    \includegraphics[height=0.065\textwidth]{Imgs/examples/ILSVRC2012_test_00020726#1} &%
    \includegraphics[height=0.065\textwidth]{Imgs/examples/sun_abhdwxcqinxtyvao#1} &%
    \includegraphics[height=0.065\textwidth]{Imgs/examples/ILSVRC2012_test_00076418#1} %
}

\CheckRmv{%
\begin{figure}[!t]
    \centering
    \footnotesize
    \renewcommand{\arraystretch}{0.2}
    \setlength{\tabcolsep}{0.3mm}
    \resizebox{\linewidth}{!}{%
    \begin{tabular}{ccccccc}
        \rotatebox[origin=l]{90}{ Image} & \AddImg{.jpg} \\ \\[0.1mm]
        \rotatebox[origin=l]{90}{\hspace{2mm} GT}& \AddImg{.png} \\ \\[0.1mm]
        \rotatebox[origin=l]{90}{Ours} & \AddImg{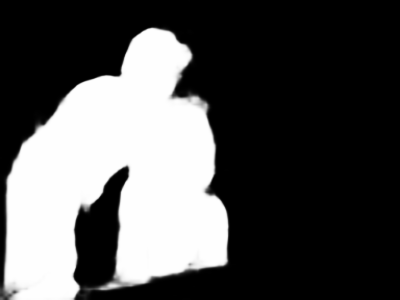} \\ \\[0.1mm]
        \rotatebox[origin=l]{90}{\hspace{-3mm} EDN-Lite} & \AddImg{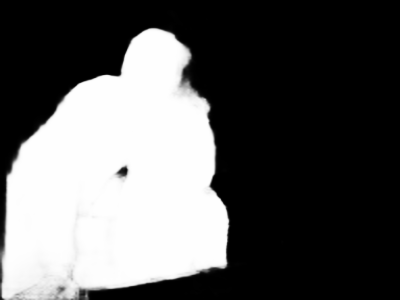} \\ \\[0.1mm]
        \rotatebox[origin=l]{90}{\hspace{-3mm} HVPNet} & \AddImg{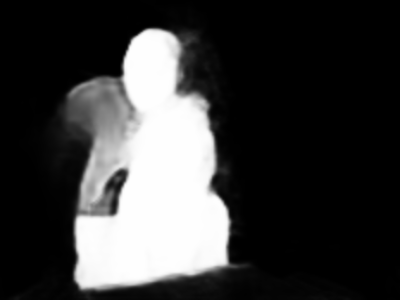} \\ \\[0.1mm]
        \rotatebox[origin=l]{90}{\hspace{-1mm} CSNet} & \AddImg{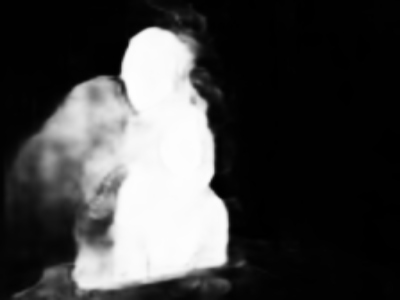} \\ \\[0.1mm]
        \rotatebox[origin=l]{90}{\hspace{-2mm} SAMNet} & \AddImg{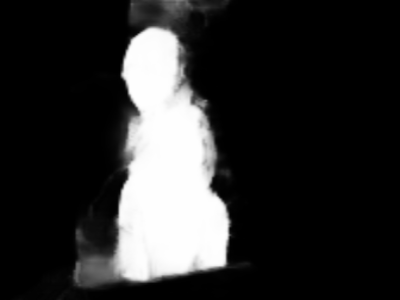} \\ \\[0.1mm]
        % & No. 1 & No. 2 & No. 3 & No. 4 \\
    \end{tabular}
    }
    \caption{\textbf{Qualitative comparison with other lightweight models on image SOD}.
    }
    % GT: ground truth.}
    \label{fig:examples}
\end{figure}%
}

\subsubsection{Qualitative comparison}
A qualitative comparison is illustrated in \figref{fig:examples}.
It is apparent that our model can accurately identify salient objects with clear boundaries and high confidence, even in complex scenarios.
Particularly in images—such as the last two in the figure—where the foreground salient object blends with the background,
competing models often incorrectly classify nearby background elements as part of the foreground.
In contrast, our model maintains precise segmentation, demonstrating its robustness and accuracy.

\subsection{Ablation Study} \label{sec:ablation}
To demonstrate the efficacy of various modules within our model, as well as the impact of different deep supervision combinations, we conducted an ablation study using the DUTS-TE dataset.
This study utilized efficiency metrics and selected performance metrics: $F_\beta^{\text{max}}$, $F_\beta^w$, and $\rm MAE$.
% As in previous section, FLOPs is computed with input image size of $320 \times 320$ and speed is tested with batch size 1 and 30.

% Impact of the output dimension of adaptive pooling layer of  {\PCASA} module.
% Table generated by Excel2LaTeX from sheet 'Ablation study'
\begin{table}[!t]
  \centering
  \caption{\textbf{Ablation study of the output dimension of the pooling layer in {\PCASA}}.}
  \resizebox{\linewidth}{!}{%
    \begin{tabular}{l|c|c|c|c|c|c}
    \Xhline{1pt}
    \multirow{2}[2]{*}{Method} & \#Param & FLOPs & Speed & \multirow{2}[2]{*}{$F_\beta^{\text{max}}$} & \multirow{2}[2]{*}{$F_\beta^w$} & \multirow{2}[2]{*}{$\rm MAE$} \bigstrut[t]\\
          & (M)   & (G)   & (FPS) &       &       &  \bigstrut[b]\\
    \Xhline{1pt}
    w/o Attn.  & 1.96  & 1.25  & 609 & 0.864 & 0.801 & 0.043 \bigstrut[t]\\
    $m=1$   & 1.99  & 1.26  & 578 & 0.864 & 0.800 & 0.043 \\
    $m=3$   & 1.99  & 1.26  & 572 & 0.865 & 0.803 & 0.043 \\
    $m=7$   & 1.99  & 1.26  & 571 & \textbf{0.867} & \textbf{0.804} & \textbf{0.042} \\
    % $m=14$  & 1.99  & 0.88  & 91 (884) & 0.867 & \textbf{0.805} & \textbf{0.042} \\
    $m=28$  & 1.99  & 1.47  & 408 & 0.865 & 0.803 & \textbf{0.042} \bigstrut[b]\\
    \Xhline{1pt}
    \end{tabular}%
  }
  \label{tab:ablation-lrsa}%
\end{table}%

\subsubsection{Attention module in {\PCASA}}
The experimental results concerning the attention module of {\PCASA} are summarized in \tabref{tab:ablation-lrsa}. It is important to note that $m$ represents the output dimension of the adaptive pooling layer in \equref{equ:lrsa}, and the attention module was evaluated with four different pooling sizes, as well as without the module for comparison. The findings indicate that the attention module enhances estimation accuracy with a minimal increase in the number of parameters, particularly when the pooling size $m$ exceeds 1. While the increase in FLOPs is generally negligible, it becomes more substantial at $m=28$.

Specifically, with $m=7$, both $F_\beta^{\text{max}}$ and $F_\beta^w$ show an improvement of 0.3\% over the model without the attention module. These accuracy enhancements are achieved with a slight increase in parameters (0.03M) and FLOPs (0.01G). Among the tested sizes, a mid-size $m=7$ offers superior performance compared to both larger ($m=28$) and smaller sizes ($m=1$ and $m=3$). Consequently, the proposed {\PCASA} with a self-attention module at $m=7$ effectively enhances estimation accuracy with a negligible efficiency trade-off.

% Impact of the split ratio of pyramid convolution module.
\begin{table}[!t]
  \centering
  \setlength\tabcolsep{9pt}
  \caption{\textbf{Effect of the split ratios of pyramid convolution module}.}
  \resizebox{\linewidth}{!}{%
    \begin{tabular}{l|c|c|c|c}
    \Xhline{1pt}
    \multirow{2}[2]{*}{Method} & \multirow{2}[2]{*}{Split Ratios} & \multirow{2}[2]{*}{$F_\beta^{\text{max}}$} & \multirow{2}[2]{*}{$F_\beta^w$} & \multirow{2}[2]{*}{$\rm MAE$} \bigstrut[t]\\
          &       &       &  \bigstrut[b]\\
    \Xhline{1pt}
    Identical Split \cite{wu2022edn}
    & [$\frac{1}{4}$, $\frac{1}{4}$, $\frac{1}{4}$, $\frac{1}{4}$] & 0.863 & 0.801 & 0.043 \bigstrut[t] \\
    Ours & [$\frac{1}{8}$, $\frac{1}{8}$, $\frac{1}{4}$, $\frac{1}{2}$] & \textbf{0.867} & \textbf{0.804} & \textbf{0.042} \bigstrut[b]\\
    \Xhline{1pt}
    \end{tabular}%
  }
  \label{tab:ablation-gpc}% 
\end{table}%

We additionally compare our model with the identical split setting.
The results of this comparison are detailed in \tabref{tab:ablation-gpc}.
Since both models exhibit equivalent efficiencies, a comparison of efficiency metrics was not conducted.
The data demonstrate that accuracy can be slightly enhanced by employing the proposed split ratios, which apply lower dilation ratios to high-scale features.

% Impact of global feature extractor.
\begin{table}[!t]
  \centering
  \caption{\textbf{Effect of the global feature extractor (GFE)}.}
  \resizebox{\linewidth}{!}{%
    \begin{tabular}{l|c|c|c|c|c|c}
    \Xhline{1pt}
    \multirow{2}[2]{*}{Method} & \#Param & FLOPs & Speed & \multirow{2}[2]{*}{$F_\beta^{\text{max}}$} & \multirow{2}[2]{*}{$F_\beta^w$} & \multirow{2}[2]{*}{$\rm MAE$} \bigstrut[t]\\
          & (M)   & (G)   & (FPS) &       &       &  \bigstrut[b]\\
    \Xhline{1pt}
    w/o GFE & 1.61  & 1.08  & 699 & 0.836 & 0.758 & 0.051 \bigstrut[t]\\
    % +ASPP  & 1.97  & 1.24  & 578 & 0.854 & 0.785 & 0.045 \\
    +ED \cite{wu2022edn}   & 1.98  & 1.21  & 580 & 0.853 & 0.783 & 0.047 \\
    Ours   & 1.99  & 1.26  & 571 & \textbf{0.867} & \textbf{0.804} & \textbf{0.042} \bigstrut[b]\\
    \Xhline{1pt}
    \end{tabular}%
  }
  \label{tab:ablation-guide}%
\end{table}%

% Impact of deep supervision combinations. F, B, C and O denote full map, boundary, center and others saliencies.
% Table generated by Excel2LaTeX from sheet 'Ablation study'
\begin{table}[!t]
  \centering
  \caption{\textbf{Ablation study of granularity-aware supervision}. F, B, C and O denote the saliency of full map, boundary, center and others, respectively. C-O indicates the combination of center and others foregrounds. Side-outputs are the intermediate representations shown in \figref{fig:our-network}.}
  \resizebox{\linewidth}{!}{%
    \begin{tabular}{c|c|c|c|c|c|c}
    \Xhline{1pt}
    \multirow{2}[3]{*}{Side-outputs} & \multicolumn{6}{c}{Setting} \bigstrut[b]\\
    \cline{2-7}          & (a)   & (b)   & (c)   & (d)   & (e)   & (f) Ours \bigstrut\\
    \Xhline{1pt}
     $D_3$ & F     & F     & F     & F     & F     & F \bigstrut[t]\\
     $D_L$ & F     & B     & B     & B     & -     & - \\
     $D_H$ & F     & C     & O     & C-O   & -     & - \\
     $G_f$ & F     & -     & -     & -     & C     & - \\
     $D_2$ & F     & C-O   & C-O   & C-O   & -     & C \\
     $D_1$ & F     & B-O   & B-O   & B-O   & B-O   & B-O \bigstrut[b]\\
    \hline
    $F_\beta^{\text{max}}$  & 0.858 & 0.848 & 0.854 & 0.855 & 0.854 & \textbf{0.867} \bigstrut[t]\\
    $F_\beta^w$ & 0.792 & 0.778 & 0.786 & 0.786 & 0.789 & \textbf{0.804} \\
    $\rm MAE$   & 0.044 & 0.048 & 0.046 & 0.045 & 0.045 & \textbf{0.042} \bigstrut[b]\\
    \Xhline{1pt}
    \end{tabular}%
  }
  \label{tab:ablation-dds}%
\end{table}%

\subsubsection{Global feature extractor} 
Subsequently, we conducted an ablation study to evaluate the impact of the global feature extractor, with the results detailed in \tabref{tab:ablation-guide}.
Our analysis compares our model, which utilizes an efficient attention mechanism for global feature extraction, against two alternatives: one without any global features and another employing extreme downsampling (ED) \cite{wu2022edn} for global feature extraction.
The results indicate that incorporating a global feature extractor significantly enhances estimation accuracy.
This improvement corroborates previous findings that global features play a crucial role in salient object detection (SOD) tasks, as highlighted in prior works \cite{chen2018reverse, wu2022edn, yun2023towards}.

Specifically, implementing ED atop the encoder to extract global features notably increases accuracy by 1.7\%. The integration of our proposed attention-based feature extractor further amplifies $F_\beta^{\text{max}}$ by an additional 1.4\%.
Such marked enhancements validate the effectiveness of attention modules in assimilating global features, affirming their utility in complex SOD tasks.

\subsubsection{Granularity-aware supervision}
Finally, we evaluated the effectiveness of various deep supervision settings within the proposed structure, and the results are summarized in \tabref{tab:ablation-dds}.
Setting (a) serves as the baseline, where both decoder side-outputs and global features are supervised using the full saliency map.
In settings (b)-(d), the decoder side-outputs are used to supervise saliencies of different granularity.
However, in settings (e) and (f), only the decoder outputs that incorporate global features are utilized for supervision.

The results indicate that supervising side-outputs with different saliency granularities does not generally enhance performance, with the exception of our method. Specifically, the high-level side-output $D_2$ is supervised using center saliency, and the low-level side-output $D_1$ is supervised using boundary-other saliency, which leads to improved performance.
In contrast, supervising the side-outputs $D_L$ and $D_H$, which do not integrate global features, does not yield performance gains.

\section{Conclusion}

In this study, we introduced \netname, a lightweight framework for both image and video SOD. 
With granularity-aware connections, the model fuses low- and high-level features under supervision signals aligned with their granularities, \ie, object locations for coarse levels and boundaries for fine levels.
%
%The side-outputs at both levels are supervised by ground truths matching their granularities: object locations for low granularity and boundaries for high granularity.
%
To enhance feature fusion within these connections, we designed granular pyramid convolution with efficient attention (GPC) and  cross-scale attention (CSA) strategies tailored to low-level and high-level fusions. 
Furthermore, a self-attention module was incorporated to capture global information, enabling precise object localization with minimal overhead.
%
%Additionally, we incorporated a self-attention module atop the encoder to capture global information, facilitating accurate object localization with negligible computational overhead. 
%
Experiments on multiple image and video benchmarks show that GAPNet establishes new\sArt performance among lightweight models, significantly narrowing the gap to heavyweight counterparts.

%\section*{Acknowledgment}

\section*{Acknowledgements}

This work was supported by A*STAR Career Development Fund under grant No. C233312006.

\section*{Declarations of Conflict of Interest}

The authors declared that they have no conflicts of interest to this work.

{\small
\bibliographystyle{IEEEtran}
\bibliography{reference}
}

% \clearpage

\bigskip

\newcommand{\AddPhoto}[1]{\includegraphics[width=.85in,clip,keepaspectratio]{#1}}

\begin{biography}{\AddPhoto{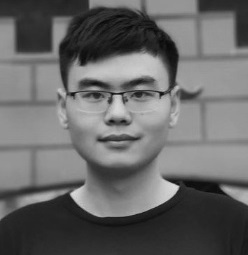}}{\textbf{Yu-Huan Wu} 
received his Ph.D. degree from Nankai University in 2022. He is a research scientist at the Institute of High Performance Computing (IHPC), A*STAR, Singapore. He has published 10+ papers on top-tier conferences and journals such as IEEE TPAMI/TIP/TNNLS/CVPR/ICCV.
His research interests include computer vision and deep learning.

E-mail: wyh.nku@gmail.com

ORCID iD: 0000-0001-8666-3435
%ORCID iD: 0000-0001-8666-3435
}
\end{biography}

\begin{biography}{\AddPhoto{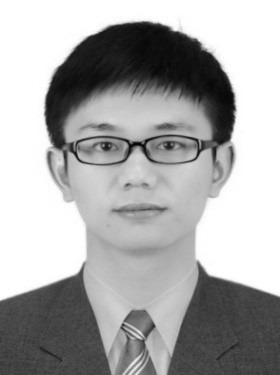}}{\textbf{Wei Liu} 
 received the bachelor’s and master’s degrees from Huazhong University of Science and Technology, China in 2015 and 2018, respectively. He then obtained the Ph.D. degree from Nanyang Technological University, Singapore in 2022. He is currently a research scientist at Institute of High Performance Computing (IHPC), Agency for Science, Technology and Research (A*STAR), Singapore. His research interests include computer vision and efficient machine learning.

 E-mail: liuw1204@gmail.com (Corresponding author)

 ORCID iD: 0000-0002-9770-8923
 }
\end{biography}

\begin{biography}{\AddPhoto{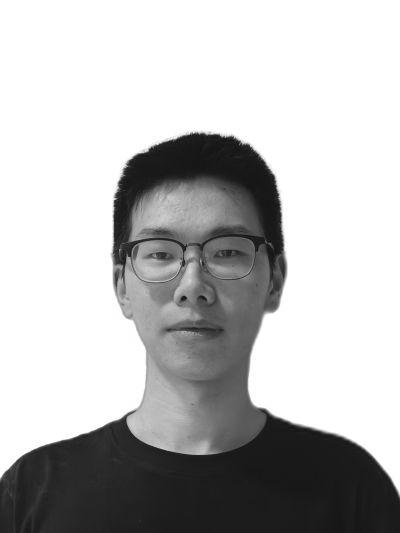}}{\textbf{Zi-Xuan Zhu} 
received his bachelor's degree from Nankai University in 2025.
He is pursuing his doctoral degree under the supervision of Prof. Deng-Ping Fan in Nankai University.
His research interests include computer vision and deep learning.

E-mail: zzxnku@mail.nankai.edu.cn

ORCID iD: 0009-0006-4357-4233
}
\end{biography}

\begin{biography}{\AddPhoto{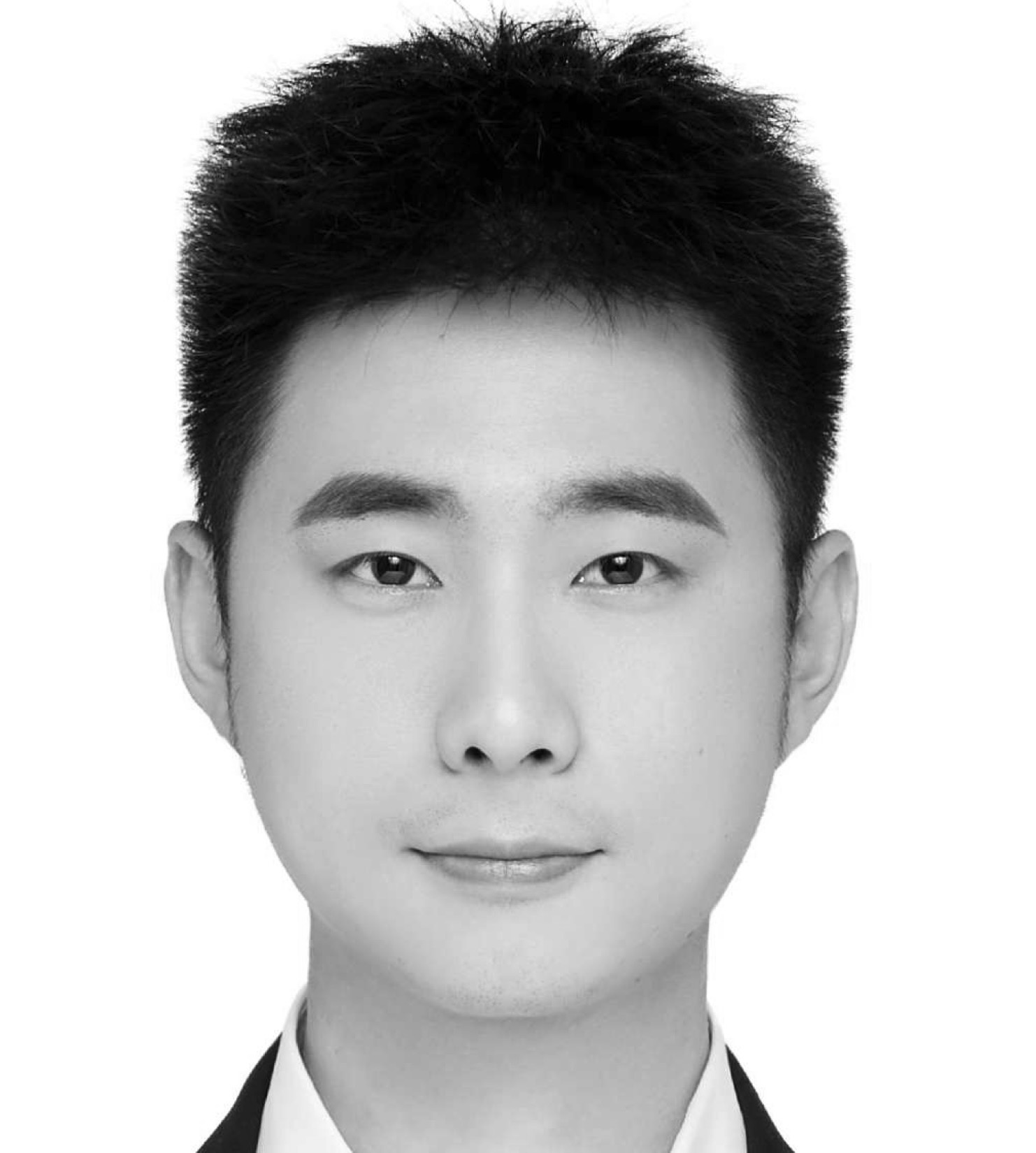}}{\textbf{Zizhou Wang} 
received his Ph.D. degree in computer science from Sichuan University in 2022. He is currently a research scientist in Institute of High Performance Computing (IHPC), Agency for Science, Technology and Research (A*STAR), Singapore. His current research interests include robust machine learning and Intelligent medical imaging.

E-mail: wang\_zizhou@ihpc.a-star.edu.sg

ORCID iD: 0000-0003-2234-9409
}
\end{biography}

\begin{biography}{\AddPhoto{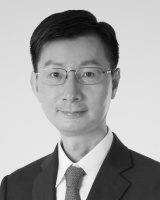}}{\textbf{Yong Liu}
is Deputy Department Director, Computing \& Intelligence Department at Institute of High Performance Computing (IHPC), A*STAR, Singapore. He is also Adjunct Associate Professor at Duke-NUS Medical School, NUS and Adjunct Principal Investigator at Singapore Eye Research Institute (SERI). 
He has led multiple research projects in multimodal machine learning, medical imaging analysis, especially AI in healthcare.

E-mail: liuyong@ihpc.a-star.edu.sg

ORCID iD: 0000-0002-1590-2029
}
\end{biography}

\begin{biography}{\AddPhoto{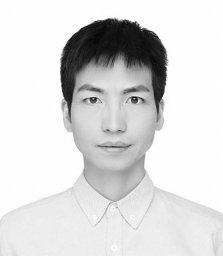}}{\textbf{Liangli Zhen}
received his Ph.D. degree from Sichuan University in 2018. He is a senior scientist and group manager at the Institute of High Performance Computing (IHPC), A*STAR, Singapore. His research interests include machine learning and optimization. He has led/co-led multiple research initiatives in robust multimodal learning. His research findings have been published in top tier journals and conferences, including IEEE TPAMI, TNNLS, ICCV, and CVPR. 

E-mail: llzhen@outlook.com (Corresponding author)

ORCID iD: 0000-0003-0481-3298
}

\end{biography}

\end{document}